\documentclass{article} 
\usepackage{iclr2021_conference,times}

\usepackage[utf8]{inputenc} 
\usepackage[T1]{fontenc}    
\usepackage{hyperref}       
\usepackage{url}            
\usepackage{booktabs}       
\usepackage{amsfonts}       
\usepackage{nicefrac}       
\usepackage{microtype}      

\usepackage{epsfig}
\usepackage{graphicx}
\usepackage{amsmath}
\usepackage{amssymb}

\usepackage{bm}
\usepackage{amsmath}
\usepackage{algorithm}
\usepackage{algorithmic}
\usepackage{amssymb}
\usepackage{booktabs}
\usepackage{subfigure}
\usepackage{textcomp}
\usepackage{multirow}
\usepackage{wrapfig}
\usepackage{caption}
\usepackage{comment}
\usepackage{amsthm}

\usepackage{comment}
\usepackage{mathrsfs}
\usepackage{makecell}
\usepackage{algorithm}
\usepackage{algorithmic}
\usepackage{color}
\usepackage{diagbox}
\usepackage{setspace}

\DeclareMathOperator*{\argmin}{\arg\min}
\DeclareMathOperator*{\Exp}{\mathbb{E}}

\newcommand{\E}{\mathcal{E}}

\newcommand{\M}{\mathcal{M}}

\newcommand{\wrt}{{{w.r.t.}}}
\newcommand{\ie}{{\emph{i.e.}}}
\newcommand{\eg}{{\emph{e.g.}}}
\newcommand{\FLOPs}{\mbox{FLOPs}}

\usepackage{array}
\newcommand{\PreserveBackslash}[1]{\let\temp=\\#1\let\\=\temp}
\newcolumntype{C}[1]{>{\PreserveBackslash\centering}p{#1}}
\newcolumntype{R}[1]{>{\PreserveBackslash\raggedleft}p{#1}}
\newcolumntype{L}[1]{>{\PreserveBackslash\raggedright}p{#1}}

\newcommand{\qileft}{[\kern-0.15em[}
\newcommand{\qiLeft}{\left[\kern-0.4em\left[}
\newcommand{\qiright}{]\kern-0.15em]}
\newcommand{\qiRight}{\right]\kern-0.4em\right]}

\newcommand{\st}{{\mbox{s.t.}}}

\renewcommand{\Roman}[1]{\uppercase\expandafter{\romannumeral#1}}
\newcommand{\red}[1]{{\color{red}{#1}}}
\newcommand{\blue}[1]{{\color{blue}{#1}}}

\renewcommand{\c}{{\bm{c}}}

\newcommand{\s}{{\bm{s}}}

\newcommand{\w}{{\bm{w}}}

\newcommand{\A}{{\mathcal{A}}}
\newcommand{\B}{\mathcal{B}}
\newcommand{\C}{\mathcal{C}}
\newcommand{\D}{\mathcal{D}}
\newcommand{\F}{\mathcal{F}}

\newcommand{\I}{\mathcal{I}}
\renewcommand{\L}{\mathcal{L}}
\newcommand{\N}{\mathcal{N}}
\renewcommand{\O}{\mathcal{O}}

\usepackage{hyperref}
\usepackage{url}

\title{Locally Free Weight Sharing for Network Width Search}

\author{Xiu Su$^{1}$, Shan You$^{2,3}$\thanks{Corresponding author.}, Tao Huang$^{2}$, Fei Wang$^{2}$, Chen Qian$^{2}$, Changshui Zhang$^{3}$, Chang Xu$^{1}$ \\
$^1$School of Computer Science, Faculty of Engineering, The University of Sydney, Australia\\
$^2$SenseTime Research\\
$^3$Institute for Artificial Intelligence, Tsinghua University (THUAI)\\
~~Beijing National Research Center for Information Science and Technology (BNRist) \\
~~Department of Automation, Tsinghua University, Beijing, P.R.China\\
~~\texttt{xisu5992@uni.sydney.edu.au, youshan@sensetime.com,} \\  
~~\texttt{\{huangtao,wangfei,qianchen\}@senseauto.com,} \\
~~\texttt{zcs@mail.tsinghua.edu.cn, c.xu@sydney.edu.au} 
}

\iclrfinalcopy 
\begin{document}

	\maketitle

	\begin{abstract}
		Searching for network width is an effective way to slim deep neural networks with hardware budgets. With this aim, a one-shot supernet is usually leveraged as a performance evaluator to rank the performance \wrt~different width. Nevertheless, current methods mainly follow a manually fixed weight sharing pattern, which is limited to distinguish the performance gap of different width. In this paper, to better evaluate each width, we propose a loCAlly FrEe weight sharing strategy (CafeNet) accordingly. In CafeNet, weights are more freely shared, and each width is jointly indicated by its base channels and free channels, where free channels are supposed to locate freely in a local zone to better represent each width. Besides, we propose to further reduce the search space by leveraging our introduced FLOPs-sensitive bins. As a result, our CafeNet can be trained stochastically and get optimized within a min-min strategy. Extensive experiments on ImageNet, CIFAR-10, CelebA and MS COCO dataset have verified our superiority comparing to other state-of-the-art baselines. For example, our method can further boost the benchmark NAS network EfficientNet-B0 by 0.41\% via searching its width more delicately.
	\end{abstract}

	\section{Introduction}
	
	Deep neural networks are easily constructed by stacking multiple layers of non-linearities on top of each other. But it is a much more delicate job than it seems. Before going into the deep, we have to determine the width of each layer\footnote{Other literature also use the number of channels/filters to indicate the network width.} \citep{na} to be stacked, especially given the consideration of the computational budgets \citep{dr,dc,bn,ms} (\eg, FLOPs, and latency) and the resulting network performance. The width of a pretrained over-parameterized neural network can be further trimmed by classical channel pruning techniques \citep{rethinkingpruning,cp,tang2019bringing} to eliminate the redundancy. Inspired by the spark of neural architecture search (NAS)  \citep{darts,once,tf,yang2020ista,huang2020explicitly,yang2021towards}, recent methods attempt to directly search for the optimal network width for a given network, and have achieved remarkable performance in various FLOPs levels or acceleration rates, such as AutoSlim \citep{autoslim}, MetaPruning \citep{metapruning} and TAS \citep{tas}. Then the obtained models can be deployed in edge devices, or even boosted by other compact techniques, such as quantization and knowledge distillation \cite{hinton2015distilling,you2017learning,kong2020learning,du2020agree}. 
	
	To determine the optimal network width, we need to evaluate and compare the performance of different network width. One of the most straightforward yet authentic approaches is to examine their training-from-scratch performance and traverse all possible settings of network width (not exceeding a maximum allowable width). However, exhaustive training is computationally unaffordable due to the huge search space. Taking MobileNetV2 as an example, 100 channels in each of the 25 layers result in $100^{25}$ possible network architectures. For sake of searching efficiency, current methods \cite{na,autoslim,tas} usually follow a weight sharing strategy by leveraging a one-shot supernet, with various sub-networks sharing the same weights with the supernet. Then each network width can be efficiently evaluated by querying the performance of its corresponding sub-network in the supernet.

\begin{figure}[t]
		\centering
		\label{motivation_1}
		\subfigure[Toy example]{
			\centering
			\label{Toy_example}
			\includegraphics[width=0.29\linewidth]{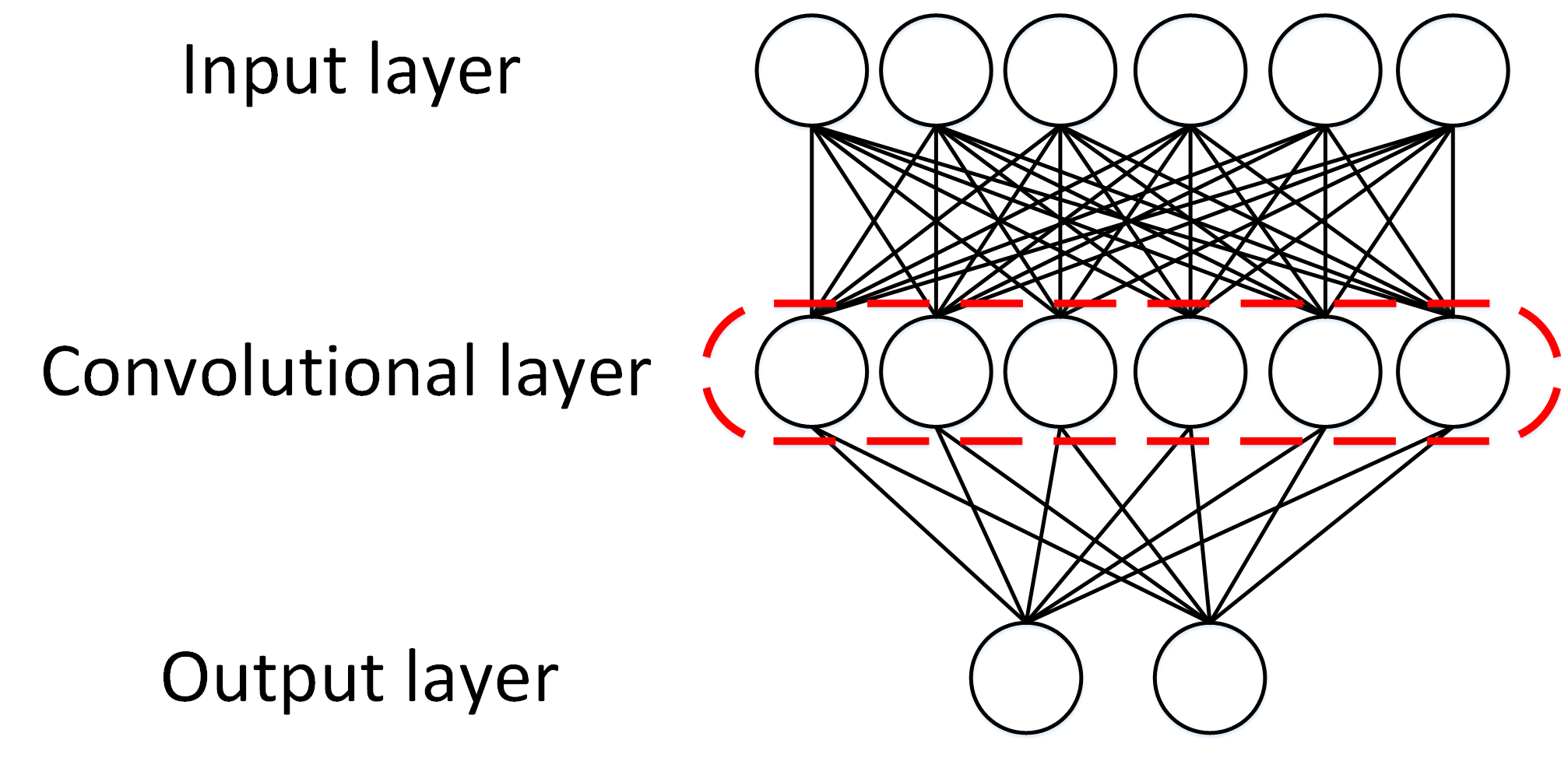}} 
		\subfigure[Channel  assignment of sub-network \wrt~differernt weight sharing patterns]{
			\centering
			\label{weight_pattern_supp}
			\includegraphics[width=0.69\linewidth]{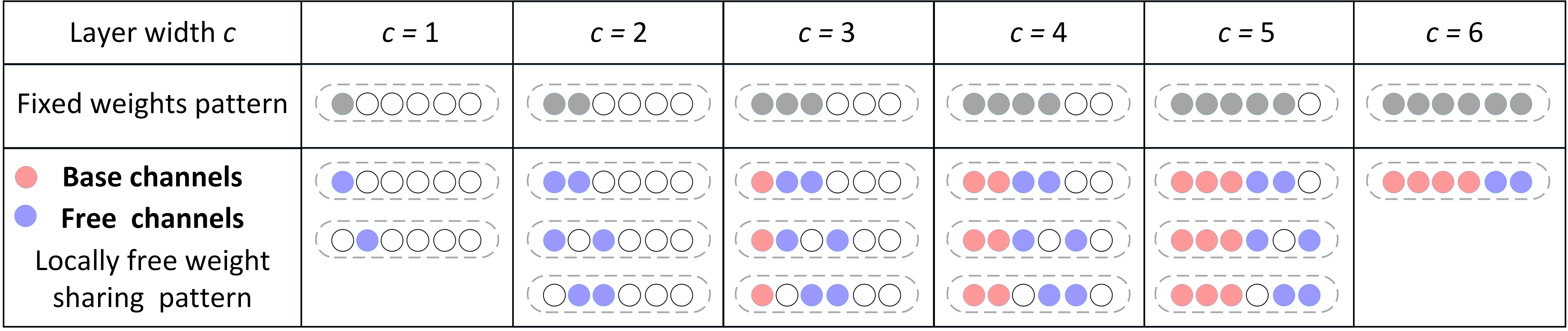}}
		\caption{(a) A toy example of the supernet. We only search for the width for its one convolutional layer with all 6 channels. (b) Examples of different weight sharing patterns. For the fixed weight sharing pattern, leftmost $c$ channels are assigned for the sub-network of width $c$. For the locally free weight sharing pattern, the offset of free zone is set to $r=1$. }
		\vspace{-6mm}
	\end{figure}

	In this way, how to specify the sub-network for each network width indicates the search space and matters for the performance evaluation.  Without loss of generality, we count channels in a layer from the left. A popular weight sharing strategy for the supernet then follows a fixed weight sharing pattern,  which simply assigns the left $c$ channels as the sub-network for the width $c$ (see the gray dots in Fig.\ref{weight_pattern_supp}). In this manner, the search space of a layer with $n$ channels is $\O(n)$. However, this fixed pattern imposes an inherent constraint on the search space. Any two sets of width fully share their weights, limiting the accurate evaluation ability of supernet and affecting the searching performance accordingly. Ideally, each channel from the supernet shall enjoy the freedom to be selected, \ie, any channel has a probability to occur in the subnetwork, no matter it locates at the leftmost or the rightmost of a layer. But the price of this \emph{full} channel freedom is the dramatic increase of the search space size to $\O(2^n)$, which affects the feasibility of width search. Instead of this unaffordable full freedom, we prefer restrictions and regularization over the search space to keep the channel freedom within an acceptable range, which benefits the efficiency of width search.

	In this paper, we propose a loCAlly FrEe weight sharing strategy of the supernet (\ie, CafeNet) for network width search. For each width $c$,
	our goal is to specify a sub-network within supernet by assigning $c$ channels, indicated by an index set $\I(c)$. To balance the weight sharing freedom and the searching efficiency, we split the index set $\I(c)$ into two parts, \ie, $c_b$ \textit{base channels} $\I_b(c)$ and $c_f$ \textit{free channels} $\I_f(c)$, with $\I(c) = \I_b(c) \cup \I_f(c)$ and $c = c_b + c_f$. (see Fig.\ref{weight_pattern_supp}). In detail, base channels follow the fixed weight sharing pattern, while free channels are encouraged to select freely from a given zone, which is the neighborhood of the $c$-th channel. In this way, each width $c$ can be better evaluated by  multiple locally free sub-subnetworks. The supernet CafeNet can be trained by optimizing the sub-network with the minimum loss for each width and search for the one with the maximum accuracy.	To further ease the burden of searching cost, we take a group of channels (\ie, \emph{bin}) as the minimal searching unit. The bin sizes are carefully set by encouraging FLOPs to be more evenly allocated over bins. We conduct extensive experiments on the benchmark ImageNet \citep{Imagenet} and CIFAR-10 \citep{cifar10} datasets to validate the superiority of our proposed CafeNet. The effectiveness of our CafeNet is also examined on the face attribute recognition task on CelebA \citep{celeba} dataset and the transferability of our searched backbone for object detection task on MS COCO dataset \citep{COCO}.

	
	\section{Problem Formulation}

	With a given network structure, the aim of the network width search is to optimize its width for each layer under certain budgets. Suppose the target network has $L$ layers, then the network width can be represented as a $L$-length tuple of layer-wise width $\c= (c_1, c_2, …,c_L)$ in which $c_i$ represents the width at the $i$-th layer. Without loss of clarity, we simply focus on searching for the optimal layer width of $c_i$. Denote the maximum allowable width at the $i$-th layer as $n_i$.  Then the number of all sub-networks in the supernet amounts to $\prod_{i=1}^L (2^{{n_i}}-1)$, which is absolutely huge. 
	Current methods mainly use weight sharing strategy to boost the searching efficiency. 
	In the typically fixed weight sharing pattern, each layer width $c_i\in [1:n_i] = \{1,2,...,n_i\}$ is indicated by one and only one sub-network (see Fig.\ref{weight_pattern_supp}), and all layers within the sub-network adopt the same weight sharing pattern, then the size of search space $\C$ is reduced to $\prod_{i=1}^L n_i$. 
	
	After specifying each width with its corresponding sub-network, we can evaluate it by examining the performance of its sub-network in the supernet with shared weights. Then the whole searching consists of two steps, \ie, supernet training and searching with supernet. Usually, the original training dataset is split into two datasets, \ie, training dataset $\D_{tr}$ and validation dataset $\D_{val}$. The one-shot supernet $\N$ with weights $W$ is trained by randomly sampling a width $\c$ and optimizing its corresponding sub-network with weights $\w_\c \subset W$,  \ie,
	\begin{equation}
	W^* = \mathop{\arg\min}_{\w_\c \subset W}~ \Exp_{\c\in U(\C)} \qiLeft \L_{train}(\w_c; \N, \c, \D_{tr})\qiRight, 
	\label{eq1}
	\end{equation}
	where $U(\C)$ is a uniform distribution of network width in search space, and $\Exp\qiLeft\cdot\qiRight$ is the expectation of random variables. After the supernet is trained, the performance (\eg, classification accuracy) of various network width is indicated by evaluating its sub-network on the trained supernet $\N^*$. And the optimal width corresponds to that of the highest performance on the validation dataset, \eg,
	\begin{equation}
	\c^* = \mathop{\arg\max}_{\c \in \C}~\mbox{Accuracy}(\w_\c^*, W^*; \N^*,\D_{val}), ~\st~\FLOPs(\c) \leq F_b,
	\label{eq3}
	\end{equation}
	where $F_b$ is a constraint on the FLOPs, here we consider FLOPs rather than latency as the hardware constraint since we are not targeting any specific hardware device like EfficientNet \cite{efficientnet} and other width search or channel pruning baselines \cite{dcp,you2020greedynas,tas,amc,tang2020reborn,su2020data}. The searching of Eq.\eqref{eq3} can be fulfilled efficiently by various algorithms, such as random or evolutionary search \citep{genetic1,greedy2,metapruning}.

	\section{CafeNet}
	\subsection{Locally free weight sharing}
	As illustrated before, a one-shot supernet $\N$ is usually trained as an evaluator to reflect the performance of different settings of network width. Assume we count channels in a layer from the left. To locate the sub-network in the supernet for width $c$ at a certain layer \footnote{Here we omit the subscript $i$ for indicating any width $c_i$ at $i$-th layer.}, conventional fixed weight sharing pattern assigns the leftmost $c$ channels as the sub-network as Fig.\ref{weight_pattern_supp}. Formally, for width $c$, its corresponding sub-network is specified by a fixed \textit{channel assignment} with an index set $\I(c)$ of channels, \ie, 
	\begin{equation}
	\mbox{fixed weight sharing pattern:} \quad \I(c) = [1:c]. 
	\end{equation}
	However, this hard assignment imposes an inherent constraint on the search space. Channel assignment of smaller width will always be a subset of that of larger width. Actually, this fixed pattern forces the weight sharing of two different width to the greatest extent. To quantify this property, we define a simple weight sharing degree $d$ of two any width $c$ and $\tilde{c}$ by examining the overlap of their channel assignment,  \ie,  
	\begin{equation}
	d(c,\tilde{c})  = \frac{|\I(c)\cap \I(\tilde{c})|}{\min(|\I(c)|,|\I(\tilde{c})|)} \in [0,1].
	\label{eqs}
	\end{equation}
	Intuitively, if the degree $d(c,\tilde{c})$ is larger, then sub-networks of width $c$ and $\tilde{c}$ will share more weights, and this implies that the supernet will enforce width $c$ and $\tilde{c}$ to have more similar performance.  And for the fixed pattern, the weight sharing degree reaches its maximum as Theorem \ref{thm1} \footnote{It holds naturally. Suppose $c\leq \tilde{c}$, then we have $\I(c)\cap \I(\tilde{c}) = \I(c)$ for fixed pattern. }. 
	\newtheorem{thm}{\bf Theorem}
	\begin{thm}\label{thm1}
		The weight sharing degree of fixed pattern will always be 1 for any two width. 
	\end{thm} 
	However, ideally, each width should have full freedom in the supernet to select its channel assignment. In this situation, sub-networks of two different layer width can be fully shared, partially shared, or even have no overlap, thus the weight sharing degree freely varies from 0 to 1. And each width can be better evaluated by some sub-networks, instead of a manually assigned sub-network. Nevertheless, this induces the search space to proliferate from the fixed $\O(n)$ to $\O(2^n)$, which is computationally unaffordable for practical search.

	Therefore, we bridge these two extreme situations, and propose a locally free weight sharing strategy. Basically, to specify a sub-network of width $c$ from a supernet for evaluation, we need to  assign $c$ channels from all $n$ channels per layer, with the channel assignment index set $\I(c)$. We split the index set $\I(c)$ into two parts, \ie, $c_b$ base channels $\I_b(c)$  and $c_f$ free channels $\I_f(c)$, with $\I(c) = \I_b(c) \cup \I_f(c)$ and $c = c_b + c_f$. In detail, for free channels, we encourage to select $c_f$ channels from a given zone, which is the neighborhood of the $c$-th channel, \ie, $\B(c;r)= [c-r:c+r]$, and $r$ is a preset allowed offset to control the search space with $c_f = r + 1$. Besides, base channels follow the fixed weight sharing pattern with $\I_b(c) = [0: c_b]$; let $\I_b(c) \cap \I_f(c) = \emptyset$, then we can have $c_b=\max(c-r-1,0)$.  In this way, in our formulation, we can specify multiple channel assignment for width $c$ to better evaluate its performance as Fig.\ref{weight_pattern_supp}, \ie, 
	\begin{equation}
	\mbox{locally free weight sharing pattern:} \quad \I(c) = [1: c_b] \cup \I_f(c),~\mbox{where}~\I_f(c)\subset [c-r:c+r] 
	\end{equation}
	According to Eq.\eqref{eqs}, the weight  sharing degree of our method is between $c_b/c \leq d(c,\tilde{c})\leq 1$. By properly setting offset $r$, we can achieve a tradeoff between the fixed pattern and the fully free pattern. Given a $r$, now our search space scales at $\O(\mathbb{C}_{2r+1}^{r+1}n)$, which is only a constant times (\eg, $\O(3n)$ for $r=1$, see Fig.\ref{weight_pattern_supp}) of fixed weight sharing search space. In this way, each width can be better represented, thus boosts the performance ranking ability of supernet.

	\textbf{Training with min-min optimization.} With the above search space, we can train our CafeNet in a stochastic setting as Eq.\eqref{eq1}, which simply samples a network width $\c$ first, and then optimizes its sub-network.  Nevertheless, in CafeNet a width is specified more freely by several sub-networks, and the performance of these sub-networks can be different. In this way, we propose to indicate the performance for each width by examining its sub-network with the best performance. In training CafeNet, instead of randomly optimizing a sub-network, we optimize the sub-network with the smallest training loss. To achieve this, we need to traverse all sub-networks for width $\c$. However, this can be efficiently fulfilled since our job is to identify a sub-network with the minimum loss, which can be quickly calculated in a feed-forward manner. Afterward, we only need to implement a backward update for the target sub-network with minimum loss. A detailed explanation of this min-min optimization is elaborated in Appendix \ref{a1}.

	\textbf{Searching with max-max selection.} After the CafeNet $\N$ is trained, we can evaluate each network width by examining its performance (\eg~classification accuracy) on the validation dataset $\D_{val}$. Similar to the training of CafeNet, for each width $\c$, we use the sub-network with the highest performance to indicate the performance. Then we search for the optimal $\c^*$ as Eq.\eqref{eq3}. Note that since searching itself is much faster than the supernet training,  the increased computational cost is subtle and acceptable in real practice. A detailed description of this max-max selection is presented in Appendix \ref{a1}.

	\subsection{Reducing search space with FLOPs-sensitive bins}
	Based on the locally free weight sharing pattern, each network width can be more flexibly indicated and then evaluated by the CafeNet. However, as previously illustrated, the size of the search space is considerably large and brings a tough challenge for the subsequent searching algorithms, such as random search and evolutionary search. In this way, current methods usually choose to reduce the search space by initializing the \textit{ minimum searching unit} as a group of channels (called \textit{bin}) instead of a single channel. Concretely, all channels at each layer are partitioned  into $K$ groups, with $(n_i/K)$ channels in each group for the $i$-th layer.

	However, this uniform partition neglects the extra information of channels within layers, \ie, kernel size and feature map resolution, and all of these factors are involved in the calculation of FLOPs. This inspires us that we should build bins based on the FLOPs. For simplicity, we encourage all bins in our search space to have the same FLOPs. Note that FLOPs is up to the layer width between two adjacent layers; for example, for convolution with resolution $H\times W$ and kernel size $K\times K$, its FLOPs can be calculated as
	\begin{equation}
	\FLOPs(c_{in}, c_{out}) = c_{in} \times c_{out} \times H \times W \times K^2,
	\label{eq5}
	\end{equation}
	where $c_{in}$ and $c_{out}$ are the width for the input and output of this layer, respectively. In this way, channels at different layers contribute differently to the overall FLOPs of the supernet. And if a channel at a layer contributes more to the FLOPs, then we can say it is more sensitive \wrt~FLOPs. This \textit{FLOPs-sensitivity} can be simply reflected by examining the real FLOPs of a single channel at a layer. Concretely,  for each channel at the $i$-th layer,  it serves as both the output channel of the $(i-1)$-th layer and input channel of the $({i+1})$-th layer. Thus the sensitivity $\varepsilon_i$ for the $i$-th layer can be represented as
	\begin{equation}
	\varepsilon_i = \FLOPs(n_{i-1}, 1) + \FLOPs(1, n_{i+1}),
	\label{eq6}
	\end{equation}
	where $n_{i-1}$ and $n_{i+1}$ are the width of the $(i-1)$-th and the $({i+1})$-th layer, respectively. 
	
	As a result, we simply encourage all bins to have the same FLOPs-sensitivity. To do so, we propose to initialize the bin size (\ie, number of channels in a bin) more adaptively. The intuition is that the larger FLOPs sensitivity a layer has, the fewer channels should be allocated in a bin, and vice versa. As a result, the bin size $b_i$ for $i$-th layer should satisfy  
	\begin{equation}
	b_i \propto 1/\varepsilon_i.
	\end{equation}
	In practice, we usually specify a minimum bin size (\ie, with maximum sensitivity $\varepsilon$) as $\beta$, then the bin size for each layer is
	\begin{equation}
	b_i = \beta \times \frac{\max_j \varepsilon_j}{\varepsilon_i}.
	\label{eq7}
	\end{equation}
	Actually, the minimum bin size $\beta$ in Eq.\eqref{eq7} controls the total number of bins, thus also determines the size of the search space. Specially, $b_i$ should be an integer and greater than or equal to 1. If $\beta$ is set to be smaller, the search space will be much larger and fine-grained since our search unit gets smaller and delicate accordingly.  We will discuss a multi-stage search strategy by controlling $\beta$ actively in Section \ref{multistage}, and present empirical investigations in Appendix \ref{a15}.

	\subsection{Further boosting performance with multi-stage search}\label{multistage}
	As previously illustrated, setting minimum bin size $\beta$ in Eq.\eqref{eq7} to be smaller will bring in a smaller searching unit, and thus a more fine-grained structure will be expected. However, this is accompanied by a larger search space and is challenging for both the ranking ability of supernet and searching performance. In this way, we can leverage a multi-stage searching strategy to avoid the search space being too large but still can search on a fine-grained level. In detail, we propose to search for the width under the multi-stage FLOPs budget, which is decayed linearly, \ie, 
	\begin{equation}
	\FLOPs({t}) \leftarrow \FLOPs(0) - (\FLOPs(0) - F_b) \times \frac{t}{T}, \quad \beta(t+1) \leftarrow \beta(t)/\alpha	\label{eq4}
	\end{equation}
	where $\FLOPs(t)$ represents the FLOPs budget in $t$-th stage and $\FLOPs(0)$ is the FLOPs of the supernet. $F_b$ is the target FLOPs after all $T$ stages as Eq.\eqref{maxmax}. Besides, at each stage, we shrink the minimum bin size $\beta$ with evolving speed $\alpha>0$. Thus the search space will evolve from the coarse-grained to the fine-grained, but with controllable size. More details are explored in Appendix \ref{a13} and \ref{a14}.  However, searching with the multi-stage strategy inevitably introduces more computation consumption. Therefore, we only implement this method in the small dataset (\ie~CIFAR-10).

\section{Experimental Results}
In this section, we perform extensive experiments to validate the effectiveness of our CafeNet. For all architectures, we use the SGD optimizer with momentum 0.9. Parameters $\beta$ in Eq.\eqref{eq7} and $\alpha$ in Eq.\eqref{eq4} are set to 1 and 2, respectively. Offset $r$ for locally free channels is set to 1 for all experiments. Detailed experimental settings are elaborated in Appendix \ref{a2}. 

We include various state-of-the-art width search methods for comparison, \eg~AutoSlim \citep{autoslim}, MetaPruning \citep{metapruning}, TAS \citep{tas}, GBN \citep{gbn} and FPGM \citep{fpgm}. Besides, for comprehensive comparison, we perform searching using both evolutionary search and random search, named as \textbf{CafeNet-E} and \textbf{CafeNet-R}, respectively. In addition, we also consider two vanilla baselines, \ie, Uniform and Random. Uniform: we shrink the width of each layer uniformly with a fixed factor to meet the FLOPs budget. Random: we randomly sample 20 networks under FLOPs constraint, and train them by 50 epochs, then we continue training the one with the highest performance and report its final result.

	\subsection{Results on ImageNet Dataset}
	We perform experiments on ResNet50 \citep{dr}, MobileNetV2 \citep{mobilenetv2} and EfficientNet-B0 \citep{efficientnet} to examine the performance of CafeNet on heavy and light models, as summarized in Table \ref{Experiments_Imagenet} and \ref{Experiments_Efficientnetb0}. In detail, ResNet50 and MobileNetV2 are searched from the original size, and for EfficientNet-B0, the 0.5$\times$ FLOPs is also searched from the original model. But the 1.0$\times$ FLOPs is from a supernet with 1.5$\times$ FLOPs from uniform width scaling. The original EfficientNet-B0, ResNet50 and MobileNetV2 has 5.3M, 25.5M and 3.5M parameters and 386M, 4.1G and 300M FLOPs with 76.4\%, 77.8\% and 73.7\% Top-1 accuracy, respectively. Note that we also adopt KD in the final training of MobileNetV2 for fair comparison with baseline methods, \ie, AutoSlim \citep{autoslim}, TAS \citep{tas}. More detailed experiments and visualization results of our searched network are shown in Appendix \ref{a8} and \ref{a10}.

	\begin{table*}[!t]
		\centering
		\scriptsize
		\caption{
			Results on ImageNet of ResNet50 and MobileNetV2 compared with state-of-the-art methods. References of baseline methods are summarized in the Appendix \ref{a4}.
		}
		\label{Experiments_Imagenet}
		\vspace{-2mm}
		\begin{tabular}{c|l|p{0.5cm}<{\centering}p{0.5cm}<{\centering}|p{0.6cm}>{\centering}p{0.6cm}<{\centering}||c|l|p{0.5cm}<{\centering}p{0.5cm}<{\centering}|p{0.6cm}<{\centering}p{0.6cm}<{\centering}}\hline
			\multicolumn{6}{c||}{ResNet50} & \multicolumn{6}{c}{MobileNetV2}\\ \hline
			&Methods&FLOPs&Param&Top-1&Top-5&&Methods&FLOPs&Param&Top-1&Top-5\\ \hline
			\multirow{8}{*}{3G} &AutoSlim & 3.0G & 23.1M & 76.0\% & -  & \multirow{7}*{200M} & MetaPruning & 217M & - & 71.2\% & -  \\
			& MetaPruning & 3.0G & - & 76.2\% & - && LEGR & 210M & - & 71.4\% & - \\
			& LEGR & 3.0G & -  & 76.2\% & - && AutoSlim & 207M & 4.1M  & 73.0\% & - \\
			& Uniform & 3.0G & 19.1M & 75.9\% & 93.0\% && Uniform & 217M & 2.7M & 70.9\% & 89.4\% \\
			& Random & 3.0G & - & 75.2\% & 92.5\% && Random & 217M & - & 70.3\% & 89.1\% \\
			& \textbf{CafeNet-R} & 3.0G & 22.6M & \textbf{77.1\%} & 94.3\% && \textbf{CafeNet-R} & 217M & 3.0M & \textbf{73.3\%} & 91.1\% \\ 
			& \textbf{CafeNet-E} & 3.0G & 23.8M & \textbf{77.4\%} & 94.5\% && \textbf{CafeNet-E} & 217M & 3.3M & \textbf{73.4\%} & 91.2\% \\ \cline{1-6} \cline{7-12}
			
			\multirow{11}*{2G} & GBN & 2.4G & 31.8M & 76.2\% & 92.8\% &\multirow{9}*{150M}& LEGR & 180M & - & 70.8\% & - \\  
			& LEGR & 2.4G & - & 75.7\% & 92.7\% && TAS & 150M & - & \textbf{70.9\%} & - \\
			& FPGM & 2.4G & - & 75.6\% & 92.6\% && AMC & 150M & - & \textbf{70.8\%} & - \\ 
			& TAS & 2.3G & - & 76.2\% & 93.1\% && LEGR & 150M & - & 69.4\% & - \\
			& AutoSlim & 2.0G & 20.6M & 75.6\% & - && MuffNet & 149M & - & 63.7\% & - \\
			& Uniform & 2.0G & 13.3M & 75.1\% & 92.7\% && Uniform& 150M & 2.0M & 69.3\% & 88.9\% \\
			& Random & 2.0G & - & 74.6\% & 92.2\% && Random & 150M & - & 68.8\% & 88.7\% \\
			& \textbf{CafeNet-R} & 2.0G & 19.1M & \textbf{76.5\%} & 93.1\% && \textbf{CafeNet-R} & 150M & 2.7M & \textbf{71.9\%} & 90.0\% \\
			& \textbf{CafeNet-E} & 2.0G & 18.4M & \textbf{76.9\%} & 93.3\% && \textbf{CafeNet-E} & 150M & 3.0M & \textbf{72.4\%} & 90.4\% \\ \cline{7-12}
			& MetaPruning & 1.0G & - & 73.4\% & -  & \multirow{11}*{100M} & MetaPruning & 105M & - & 65.0\% & - \\
			& AutoSlim & 1.0G & - & 74.0\% & - && Uniform & 105M & 1.5M & 65.1\% & 89.6\% \\ \cline{1-6}

			\multirow{4}*{1G}& Uniform & 1.0G & 6.6M& 73.1\% & 91.8\% && Random & 105M & - & 63.9\% & - \\
			& Random & 1.0G & - & 72.2\% & 91.4\% && \textbf{CafeNet-R} & 106M & 2.2M & \textbf{68.2\%} & 88.2\% \\  
			& \textbf{CafeNet-R} & 1.0G & 11.2M & \textbf{74.9\%} & 92.3\% && \textbf{CafeNet-E} & 106M & 2.1M & \textbf{68.7\%} & 88.5\% \\
			& \textbf{CafeNet-E} & 1.0G & 12M & \textbf{75.3\%} & 92.6\% && MuffNet & 50M & - & 50.3\% & -\\ \cline{1-6}
			
			\multirow{5}*{570M} & AutoSlim & 570M & - & 72.2\% & - && MetaPruning & 43M & - & 58.3\% & - \\
			& Uniform & 570M & 4.0M & 71.6\% & 90.6\% && Uniform & 50M & 0.9M & 59.7\% & 82.0\% \\
			& Random & 570M & - & 69.4\% & 90.3\% && Random & 50M & - & 57.4\% & 81.2\% \\ 
			& \textbf{CafeNet-R} & 570M & 11.3M & \textbf{72.7\%} & 90.9\% && \textbf{CafeNet-R} & 50M & 1.7M & \textbf{64.3\%} & 85.2\% \\ 
			& \textbf{CafeNet-E} & 570M & 12.0M & \textbf{73.3\%} & 91.2\%&& \textbf{CafeNet-E} & 50M & 1.6M & \textbf{64.9\%} & 85.4\% \\ \hline \hline
		\end{tabular}	
		\vspace{-2mm}
	\end{table*}

	\begin{table*}[!t]
		\centering
		\scriptsize
		\caption{Searching results of EfficientNet-B0 with 1$\times$ and 0.5$\times$ FLOPs on ImageNet dataset.}
		\label{Experiments_Efficientnetb0}
		\vspace{-2mm}
		{\begin{tabular}{l|c|c|c|c||l|c|c|c|c}
				\hline
				\multicolumn{5}{c||}{1$\times$  EfficientNet-B0}& \multicolumn{5}{c}{0.5$\times$ EfficientNet-B0}  \\ \hline
				Methods&FLOPs&Param&Top-1&Top-5&Methods&FLOPs&Param&Top-1&Top-5  \\ \hline
				Uniform & 385M & 5.3M & 76.42\% & 92.24\% & Uniform & 192M & 2.8M & 74.16\% & 91.48\% \\
				Random & 385M & 5.1M & 75.96\% & 92.11\% & Random & 192M & 3.0M & 73.36\% & 91.03\% \\
				\textbf{CafeNet-R} & 385M & 6.2M & \textbf{76.59\%}& 92.74\% & \textbf{CafeNet-R} & 192M & 3.7M & \textbf{74.47\%} & 91.65\% \\ 
				\textbf{CafeNet-E} & 385M & 6.9M & \textbf{76.83\%} & 92.96\% & \textbf{CafeNet-E} & 192M & 3.9M & \textbf{74.62\%} & 91.87\% \\ \hline
				
		\end{tabular}}	
		\vspace{-2mm}
	\end{table*}

	As shown in Table \ref{Experiments_Imagenet}, our CafeNet achieves the highest performance on ResNet50 and  MobileNetV2 w.r.t. different FLOPs, which proves the superiority of our method compared to other width search algorithms. Besides, CafeNet obtains remarkable performance with tiny FLOPs budgets. For example, our 570M ResNet50 achieves 73.3\% Top-1 accuracy, surpassing AutoSlim \citep{autoslim} by 1.1\%. In addition, as shown in Table \ref{Experiments_Efficientnetb0}, for the NAS-based network EfficientNet-B0, the performance can be further improved using CafeNet. It indicates that our CafeNet can optimize its width in a more fine-grained manner.

	\subsection{Results on CIFAR-10 Dataset}
	We also investigate CafeNet with MobileNetV2 \citep{mobilenetv2} and VGGNet \citep{vgg} on the moderate CIFAR-10 dataset. Our original VGGNet (MobileNetV2) has 20M (2.2M) parameters and 399M (297M) FLOPs with accuracy of 93.99\% (94.81\%). As shown in Table \ref{Experiments_CIFAR10}, our CafeNet enjoys significant superiority to other algorithms. In detail, our 44M MobileNetV2 achieves 95.31\% accuracy and outperforms MuffNet \citep{muffnet} by 2.2\%. For VGGNet, our CafeNet achieves 94.36\% accuracy with 189M FLOPs, which surpasses DCP \citep{dcp} by 0.2\%. Similarly, our CafeNet shows great advantages \wrt~tiny models, \eg~with 2.11\% accuracy improvement than AutoSlim \citep{autoslim} on MobileNetV2 with 28M FLOPs. More experiments on CIFAR-10 dataset are investigated in Appendix \ref{a9}.
	
	\begin{table*}[!t]
		\centering
		\scriptsize
		\caption{
			Results of MobileNetV2 and VGGNet on CIFAR-10 dataset. References of baseline methods are summarized in Appendix \ref{a4}.}
		\label{Experiments_CIFAR10}
		\vspace{-2mm}
		{\begin{tabular}{c|l|cc|c||c|l|cc|c}
				\hline
				\multicolumn{5}{c||}{MobileNetV2} & \multicolumn{5}{c}{VGGNet} \\ \hline
				Groups&Methods&FLOPs&Parameters&Accuracy&Groups&Methods&FLOPs&Parameters&Accuracy  \\ \hline
				\multirow{6}*{100M+} & DCP & 218M & 1.7M & 94.75\% &\multirow{5}*{200M} & GAL & 190M & - & 93.80\% \\
				& \textbf{CafeNet-R}  & 188M & 1.4M & \textbf{95.44\%} && DCP & 199M & 10.4M & 94.16\% \\
				& \textbf{CafeNet-E} & 188M  & 1.5M & \textbf{95.56\%} && Sliming & 199M & 10.4M & 93.80\% \\ 
				
				& MuffNet & 175M & - & 94.71\% && \textbf{CafeNet-R} & 189M & 8.3M & \textbf{94.27\%} \\ 
				& \textbf{CafeNet-R} & 144M & 1.2M & \textbf{95.28\%} && \textbf{CafeNet-E} & 189M & 8.0M & \textbf{94.36\%} \\ \cline{6-10}
				& \textbf{CafeNet-E} & 144M & 1.1M & \textbf{95.44\%} &\multirow{6}*{100M+} & PS & 156M & - & 93.63\% \\ \cline{1-5}
				
				\multirow{8}*{100M-} & AutoSlim & 88M & 1.5M & 93.20\% && \textbf{CafeNet-R} & 154M & 3.4M & \textbf{94.09\%} \\
				& AutoSlim & 59M & 0.7M & 93.00\% && \textbf{CafeNet-E} & 154M & 3.1M & \textbf{94.23\%} \\
				& MuffNet & 45M & - & 93.12\% && AOFP & 124M & - & 93.84\% \\	
				& \textbf{CafeNet-R} & 44M & 0.4M & \textbf{95.16\%} && \textbf{CafeNet-R} & 115M & 2.4M & \textbf{93.87\%} \\  
				& \textbf{CafeNet-E} & 44M & 0.4M & \textbf{95.31\%} && \textbf{CafeNet-E} & 115M & 2.1M & \textbf{94.01\%} \\	 \cline{6-10}
				
				& AutoSlim & 28M & 0.3M & 92.00\% & \multirow{3}*{76M} & CGNets & 92M & - & 92.88\% \\
				& \textbf{CafeNet-R} & 28M & 0.2M & \textbf{93.87\%} && \textbf{CafeNet-R} & 76M & 2.2M & \textbf{93.36\%} \\ 
				& \textbf{CafeNet-E} & 28M & 0.2M & \textbf{94.11\%} && \textbf{CafeNet-E} & 76M & 1.4M & \textbf{93.67\%} \\ \hline
				
		\end{tabular}}	
		\vspace{-3mm}
	\end{table*}

	\subsection{Experiments on face attribute task with CelebA dataset}
	To futher investigate the effectiveness of CafeNet, we perform experiments of multi-label classification task on CelebA \citep{celeba} dataset with MobileNetV2 \citep{mobilenetv2} and ResNet18 \citep{dr}. CelebA dataset has 162K training and 20K testing images with 40 attribute labels from 2 categories. We report the average accuracy of all 40 labels. The original MobileNetV2 (ResNet18) has 213M (1.3G) FLOPs and 2.2M (11.2M) parameters with 92.07\% (92.13\%) average accuracy.


	As shown in Table \ref{Experiments_CelebA}, our CafeNet can even achieve higher accuracy with FLOPs reducing a bit. For example, our 162M-FLOPs MobileNetV2 has 92.19\% average accuracy and outperforms the original model by 0.12\%. Besides, our algorithm still enjoys great superiority in tiny FLOPs; for example, our 130M-ResNet18 achieves 91.92\% average accuracy, which is 0.26\% higher than the uniform baseline. Besides, to comprehensively compare the classification performances on all 40 labels, we present the comparison results for each label in Appendix \ref{a9}.

	\begin{table*}[!t]
		\centering
		\scriptsize
		\caption{Results of MobileNetV2 and ResNet18 on CelebA dataset.}
		\label{Experiments_CelebA}
		\vspace{-2mm}
		{\begin{tabular}{l|c|c|c|c||l|c|c|c|c}
				\hline
				\multicolumn{5}{c||}{MobileNetV2} & \multicolumn{5}{c}{ResNet18} \\ \hline
				FLOPs&21M&51M&106M&162M&FLOPs&130M&316M&619M&1G  \\ \hline
				Uniform & 91.63\% & 91.73\% & 91.92\% & 91.97\% & Uniform & 91.66\% & 91.79\% & 91.93\% & 92.03\% \\
				Random & 91.42\% & 91.52\% & 91.63\% & 91.76\% & Random & 91.51\% & 91.62\% & 91.67\% & 91.84\% \\
				\textbf{CafeNet-R} & \textbf{91.71\%} & \textbf{92.03\%} & \textbf{92.09\%}& \textbf{92.12\%} & \textbf{CafeNet-R} & \textbf{91.83\%} & \textbf{92.07\%} & \textbf{92.13\%} & \textbf{92.17\%} \\ 
				\textbf{CafeNet-E} & \textbf{91.85\%} & \textbf{92.13\%} & \textbf{92.16\%} & \textbf{92.19\%} & \textbf{CafeNet-E} & \textbf{91.92\%} & \textbf{92.16\%} & \textbf{92.18\%} & \textbf{92.25\%} \\ \hline
				
		\end{tabular}}	
		\vspace{-2mm}
	\end{table*}

	\subsection{Transferability of the searched width to object detection task}
	For object detection tasks, its backbone is usually initialized by a pretrained model on ImgeNet dataset. In this way, we use MobileNetV2 and ResNet50 with 0.5$\times$ FLOPs as the backbone feature extractors to examine the transferability of our searched network width. Besides, we leverage both the two-stage Faster R-CNN with Feature Pyramid Networks (FPN) \citep{fpn,frcnn} and the one-stage RetinaNet \citep{retinanet} frameworks for verification. We train models using  the \textit{trainval} split of MS COCO \citep{COCO} as training data and report the results in mean Average Precision \citep{fpn,frcnn} on \textit{minival} split. As shown in Table \ref{Experiments_COCO}, the backbones obtained by our method (CafeNet 0.5$\times$) can achieve higher performance compared to the uniform baseline (Uni 0.5$\times$).

	\begin{table*}[!t]
		\centering
		\scriptsize
		\caption{Results of ResNet50 and MobileNetV2 on MS COCO dataset \citep{retinanet,fpn}.}
		\label{Experiments_COCO}
		\vspace{-2mm}
		{\begin{tabular}{l|c|c|c||l|c|c|c}
				\hline
				\multicolumn{4}{c||}{ResNet50} & \multicolumn{4}{c}{MobileNetV2} \\ \hline
				Framework&Original 1.0$\times$&CafeNet 0.5$\times$&Uni 0.5$\times$&Framework&Original 1.0$\times$&CafeNet 0.5$\times$&Uni 0.5$\times$  \\ \hline
				RetinaNet  & 36.3\% & 34.9\% & 34.2\% & RetinaNet & 31.2\% & 29.2\% & 27.5\% \\
				Faster R-CNN & 37.2\% & 36.0\% & 35.5\% & Faster R-CNN & 31.7\% & 29.1\% & 28.5\% \\ \hline
				
		\end{tabular}}	
		\vspace{-2mm}
	\end{table*}
	
	\subsection{Ablation Studies}

\begin{figure}[t]
	\centering
	\subfigure[Performance of CafeNet under different offsets $r$]{
		\label{fig:subfig:b} 
		\includegraphics[width=0.49\linewidth]{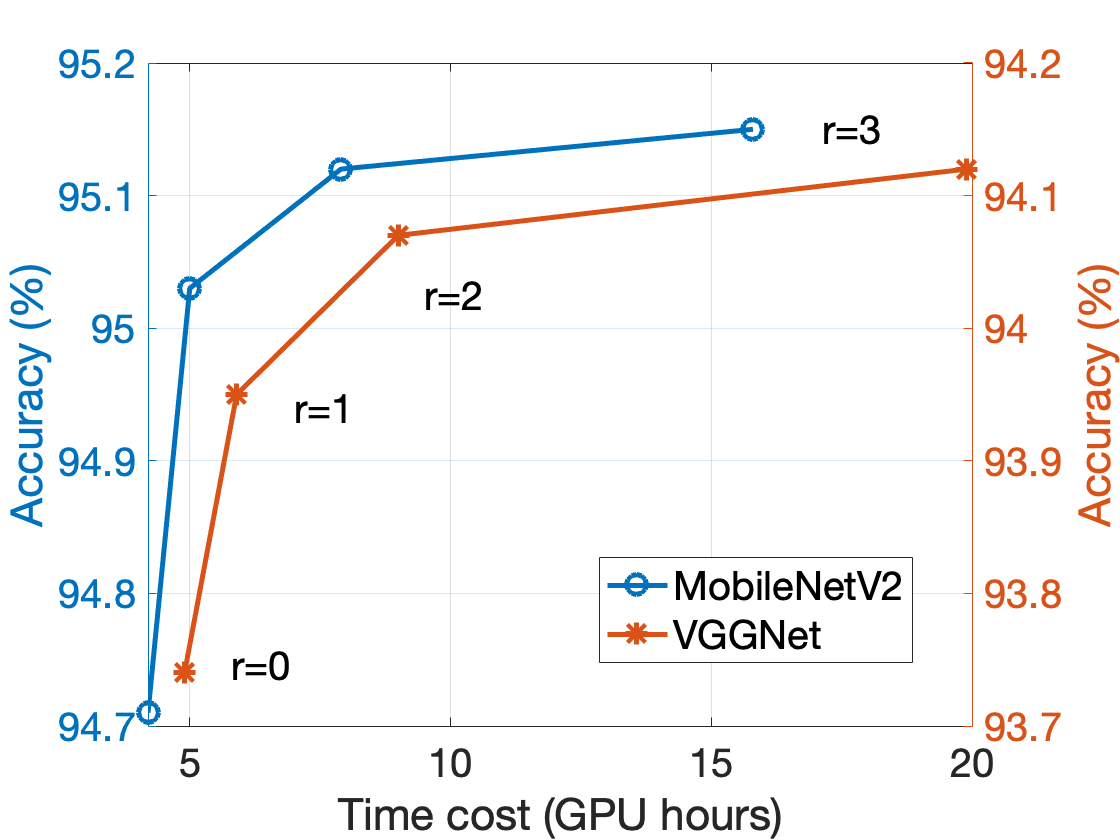}}
	~
	\subfigure[Ablations of min-min optimization]{
		\label{fig:subfig:bb}
		\includegraphics[width=0.45\linewidth]{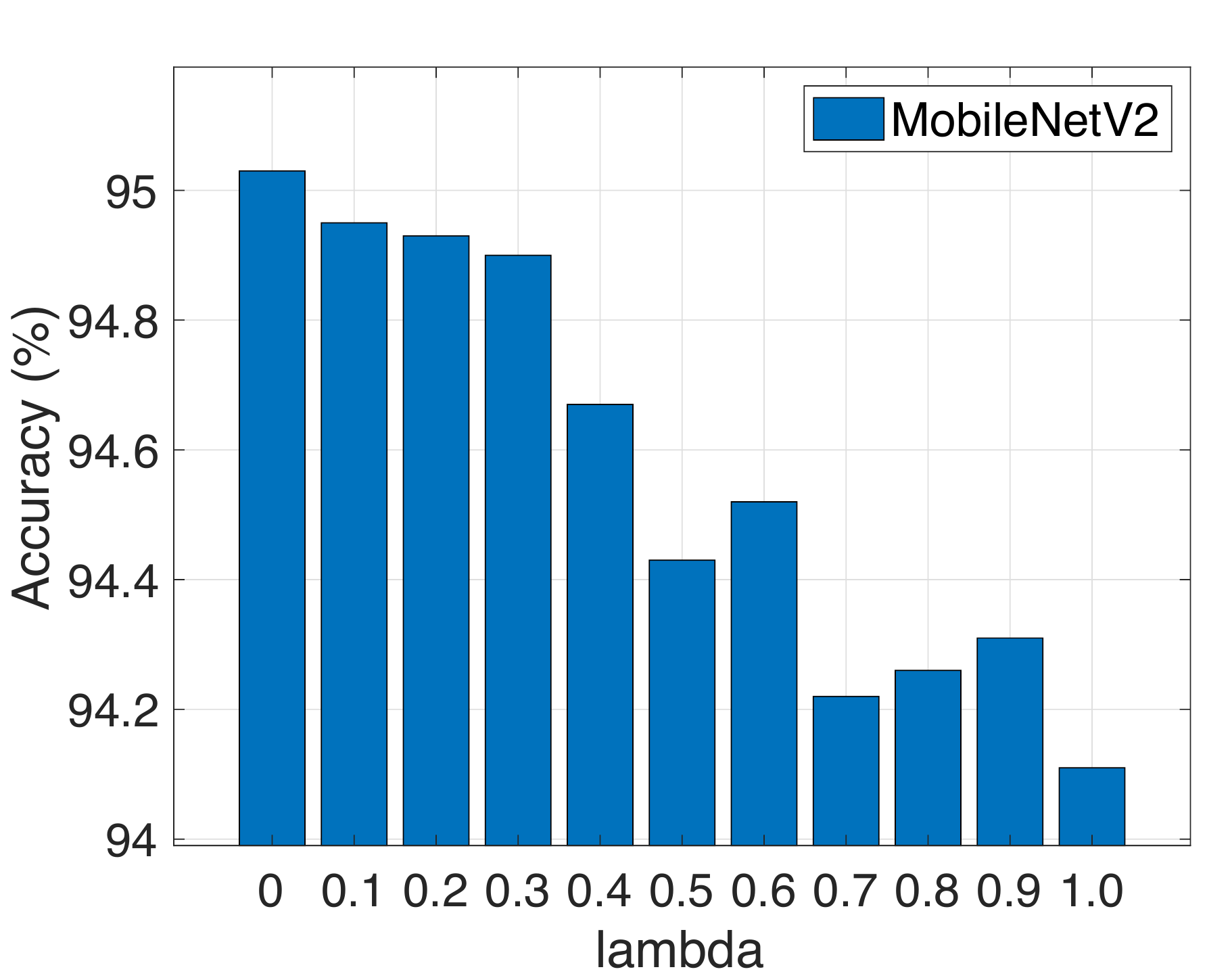}}
	\vspace{-3mm}
	\caption{(a) Performance with different offset $r$ of free channels for the 0.5$\times$-FLOPs MobileNetV2 and VGGNet on CIFAR-10 dataset. (b) Ablations of min-min optimization with 0.5$\times$-FLOPs MobileNetV2 on CIFAR-10 dataset.}
	\label{local_freedom}
	\vspace{-5mm}
\end{figure}

\textbf{Effect of offset $r$ for free channels.} As described before, free channels freely locate in the local zone indicated by an offset $r$. Since we search on bins, $r$ controls the size of search space; larger $r$ implies more accurate evaluation of width but induces larger search space, corresponding to more computational cost. To investigate its effect, we compare the performance and time cost between different $r$ on 0.5$\times$-FLOPs MobileNetV2 and VGGNet on CIFAR-10 dataset. As shown in Fig.\ref{fig:subfig:b}, accuracy performance benefits from the increase of $r$ while the larger $r$ also aggravates the burden of training. For a trade-off between performance and time cost, we set $r=1$ in all experiments. Note that in the training of CafeNet, larger $r$ will have more sub-networks; however, identifying the one with the minimum loss only involves feed-forward calculation and does not increase training cost dramatically. More results refer to Appendix \ref{a5}.

\textbf{Effect of min-min optimization on supernet}. During training CafeNet, we only optimize the sub-network with minimum training loss for each sampled width. 
To investigate the effect of this optimization strategy, suppose we have all $\tau$ iterations, then we implement min-min optimization only on the last $(1-\lambda)\cdot\tau$ iterations with $\lambda\in[0,1]$. For the first $\lambda\cdot\tau$ iterations, we simply optimize one sub-network randomly for each sampled width. As shown in Fig.\ref{fig:subfig:bb}, our 0.5$\times$-FLOPs MobileNetV2 on CIFAR-10 improves 0.92\% accuracy from $\lambda = 1$ to $\lambda = 0$, which means that min-min optimization does help to better evaluate each width and boost the searching performance accordingly. We further record the performance of 1K sampled width; details refer to Appendix \ref{a12}.


\textbf{Effect of performance improvement with more training and searching cost}. Since CafeNet needs extra training and searching cost than the fixed weight sharing pattern, one natural question comes to \textit{whether a fixed pattern can benefit from more training and searching cost and even surpass CafeNet as a result}. With this aim, we implement a fixed pattern (\ie, $r=0$ for CafeNet) with more times (1$\times$$\sim$3$\times$) of training epochs and generations using evolutionary search to search for a 0.5$\times$-FLOPs MobileNetV2 and VGGNet on CIFAR-10 dataset.

\begin{table*}[h]
	\centering
	\scriptsize
	\caption{Performance with more training and searching cost on CIFAR-10 dataset. Note CafeNet has 95.44\% and 94.36\% accuracy for 0.5$\times$-FLOPs MobileNetV2 and VGGNet, respectively.}
	\label{time_efficiency}
	\vspace{-2mm}
	\begin{tabular}{c|c|c|c||c|c|c|c}
		\hline
		\multicolumn{4}{c||}{MobileNetV2} & \multicolumn{4}{c}{VGGNet} \\ \hline
		\diagbox{Searching}{Training} &1$\times$&2$\times$&3$\times$&\diagbox{Searching}{Training} &1$\times$&2$\times$&3$\times$\\ \hline
		1$\times$&94.53\%& 94.57\% & 94.55\% &1$\times$& 93.62\%& 93.65\% & 93.66\% \\ \hline
		2$\times$&94.61\%& 94.67\% & 94.69\% &2$\times$& 93.68\% & 93.73\% & 93.71\% \\ \hline
		3$\times$&94.72\%& 94.75\% & 94.74\% &3$\times$& 93.69\% & 93.66\% & 93.74\% \\ \hline
	\end{tabular}	
	\vspace{-2mm}
\end{table*} 

As shown in Table \ref{time_efficiency}, we can see that searching with more generations by evolutionary search can only improve the searching results slightly, while more times of training cost for supernet has almost no effect on the searching performance. By comparing with Table \ref{Experiments_CIFAR10} and Fig.\ref{fig:subfig:b}, our CafeNet does efficiently improve the performance of searching results with the same or a little extra training and searching cost.

\section{Conclusion}
In this paper, we introduce a new weight sharing pattern for network width search. In detail, our locally free pattern enables each width is jointly indicated by its base channels and free channels, where free channels are used to better and more flexibly to evaluate the performance. Besides, we leverage FLOPs-sensitive bins to reduce the search space and allow FLOPs to distribute more evenly for different layers. CafeNet can be trained stochastically using min-min optimization and searches the optimal width by max-max selection.   Extensive experiments have been conducted on benchmark ImageNet, CIFAR-10, CelebA, and COCO datasets to show the superiority of our CafeNet to other state-of-the-art methods.

\subsubsection*{Acknowledgments}
This work is funded by the National Key Research and Development Program of China (No. 2018AAA0100701) and the NSFC 61876095. Chang Xu was supported in part by the Australian Research Council under Projects DE180101438 and DP210101859.	
	
\bibliography{iclr2021_conference}
\bibliographystyle{iclr2021_conference}

	\newpage 
	\appendix
	\section{Appendix}
	
	We organize our appendix as follows.  We explain the details of using CafeNet for training and searching in appendix \ref{a1}. In appendix \ref{a2}, we give details of our experiments \wrt~different datasets. We report the implementation details of our CafeNet with Algorithm 1 in appendix \ref{a3}. Then, we introduce the corresponding relationship between names and references in appendix \ref{a4}. And we explore the efficiency of CafeNet in training one-shot supernet $\N$ \ref{a5}.  We show the performance gain of each part of CafeNet in appendix \ref{a6}. We examine the effect of FLOPs-sensitive bins in appendix \ref{a7}.  We explore more detailed experiments of ImageNet, CIFAR-10, and CelebA dataset in appendix \ref{a8} and \ref{a9}. Then we show the visualization of searched network width in appendix \ref{a10}. In appendix \ref{a11}, we visualize the configuration of bins with EfficientNet-B0 and MobileNetV2.  In addition, we investigate the effect of min-min optimization with MobileNetV2 and ResNet50 in appendix \ref{a12}. We study the impact of the multi-stage search with evolving bins on CIFAR-10 dataset in appendix \ref{a13}. We provide the estimation of the size of the search space in the multi-stage search in appendix \ref{a14}. Finally, the effect of the smallest bin size $\beta$ in FLOPs-sensitive bins is shown in appendix \ref{a15}.
	
	\subsection{Training and searching with CafeNet}   \label{a1}
	
	\textbf{Training with min-min optimization.} With the bin-based search space, we can train our CafeNet in a stochastic setting as Eq.\eqref{eq1}, which simply samples a network width $\c$ first, and then optimizes the corresponding sub-network. Nevertheless, in CafeNet, a width is specified more freely by several sub-networks, and the performance of these sub-networks can be different to a great extent. According to this, we propose to indicate the performance of width by examining its sub-network with the best performance. Concretely, during the training of CafeNet, instead of randomly optimizing a sub-network, we optimize the sub-network with the smallest training loss. 
	For a network width $\c$, we denote its corresponding sub-network set as $\mathbb{S}_\c$, thus for the CafeNet $\N$ with weights $W$, the training target of min-min optimization can be written as, 
		\begin{align}
		& W^* = \mathop{\arg\min}_{W}~ \Exp_{\c\in U(\C)} \qiLeft  \L_{train}(\w_{\s_\c^*}; \N, \c, \D_{tr})\qiRight
		\label{minmin0}\\
		\st &~s_\c^*  = \argmin_{s \in \mathbb{S}_\c}~\L_{train}(\w_s; \N, \c, \D_{tr}) ,
		\label{minmin}
		\end{align}
		where $w_s \subset W$, denotes the weights of subnet $s$. To find the optimal sub-network $s_\c^*$, it needs to traverse all sub-networks and results in additional computational cost. However, this can be efficiently implemented by calculating the loss of all networks without backpropagation (\ie, \textit{no\_grad} mode in PyTorch). And then, we only need to backward once using the target sub-network $s_\c^*$.

	\textbf{Searching with max-max selection.} After the CafeNet $\N$ is trained, we can evaluate each network width by examining its performance (\eg~classification accuracy) on the validation dataset $\D_{val}$. Similar to the training of CafeNet, for a network width $\c$, we use the sub-network with the highest performance to indicate its performance. Then the searching amounts to a max-max selection problem
	\begin{equation}
	\c^* = \mathop{\arg\max}_{\c \in \C}~\max_{s \in \mathbb{S}_\c} \mbox{Accuracy}(\w_s^*, W^*; \N^*,\D_{val}), ~\st~\FLOPs(\c) \leq F_b.
	\label{maxmax}
	\end{equation}
	Note that the inner $\max$ also needs to calculate the validation accuracy of all sub-networks. Nevertheless,  since searching itself is much faster than the supernet training,  the increased computational cost is subtle and acceptable in real practice. 

	\subsection{Details of experimental settings}   \label{a2}
	
	In this section, we present the implementation details of our CafeNet \wrt~experiments on various datasets. In general, for most networks except EfficientNet-B0,  we use a SGD optimizer with momentum 0.9. While for EfficientNet-B0, we use RMSprop optimizer with epsilon set to 0.001. The parameters of $\beta$ and $\alpha$ are initialized to 1 and 2, respectively, to determine the search space. For evolutionary search,  we set the population and iteration size to 40 and 50, respectively. Besides, 2k network width are sampled for a random search. After searching, the optimized network width is trained from scratch for evaluation.
	
	\textbf{ImageNet and CIFAR-10 dataset.} we set weight decay to $10^{-4}$ for ResNet50 (same setting is also adopted for ResNet34 and ResNet18) and VGGNet while $5 \times 10^{-5}$  for MobileNetV2. The learning rate is decayed with cosine strategy from 0.1 to $10^{-5}$ for MobileNetV2 and ResNet50 with 300 training epochs and a mini-batch size of 256. Besides, for ResNet50, we adopt the strategy of random erasing with a probability of 0.4. For VGGNet, we train for 400 epochs with batch size 128, and the learning rate is decayed by step strategy from 0.1 and divided by 10 at 150-th, 225-th, 300-th epoch. For EfficientNet-B0, the learning rate is initialized to 0.13 and decayed by 0.96 for every 3 epochs. We train EfficientNet-B0 for 300 epochs, of which the first 5 are warm-up epochs, and the weight decay is set to $1 \times 10^{-5}$. 
	
	\textbf{CelebA dataset.} We set the training epochs to 15 (10) for MobileNetV2 and ResNet18 with learning rate annealed from 0.1 to $10^{-5}$ by cosine strategy, and the batch size of these two networks are set to 64.
	
	\textbf{MS COCO dataset.} We conduct object detection experiments using two popular frameworks Faster R-CNN with Feature Pyramid Networks (FPN) \citep{fpn} and RetinaNet \citep{retinanet} on MS COCO dataset \citep{COCO}. The \textit{trainval35k} split is used for training and we report the mean Average Precision (mAP) on \textit{minival} split. We train all the models using SGD for 12 epochs from ImageNet pretrained weights, and the initial learning rate is set to 0.08 with batch size 256, which decays 0.1 at 8-th and 11-th epoch.

	\subsection{Algorithm of CafeNet}   \label{a3}
	
	The details about CafeNet are presented in Algorithm 1. In specific, we search network width on CIFAR-10 dataset with 3 stages while others with 1 stage. 
	\begin{algorithm}[h]
		\caption{Locally free weight sharing for network width search}
		\label{alg}
		\begin{algorithmic}[1]
			\REQUIRE{The number of multi-stage $\M$. The training epochs of supernet $\N$. The smallest bin size $\beta$ and the bin evloving speed $\alpha$. Searching methods $\A$. Training dataset $\D_{tr}$. Validation dataset $\D_{val}$.}
			\WHILE{stage $<\M$}
			\STATE initialize the supernet $\N$ and bins groups.
			\WHILE{epochs $<\E$}
			\STATE randomly assign the training and backwards path 
			\STATE updating the corresponding path of bin groups with dataset $\D_{tr}$.
			\ENDWHILE
			\STATE update $\beta = \beta$ $/$ ${\alpha}$
			\ENDWHILE
			\STATE search the optimized network width within supernet $\N$ by searching methods $\A$ (as illustrated in section 3.3).
			\STATE train the optimized network width from scratch for evaluation.
			\ENSURE{the optimized network width with evaluation results}
		\end{algorithmic}
	\end{algorithm}

	\subsection{Index and literature}   \label{a4}

	In order to simplify the correspondence between names and references in all tables, we have listed the names used and their corresponding papers as follows:

			\begin{table*}[h]
			\centering
			\small
			\scriptsize
			\caption{The names used in tables and their corresponding papers.}
			\label{Indea_paper}
			\vspace{-2mm}
			\begin{tabular}{|lc|lc|lc|}
					\hline
					Names&papers&Names&papers&Names&papers  \\ \hline
					ResNet & \citep{dr} & AutoSlim & \citep{autoslim} & Rethinking & \citep{rethinkingpruning} \\
					MetaPruning & \citep{metapruning} & MIL & \citep{MIL} & LEGR & \citep{legr} \\
					PF & \citep{PF}	& GBN& \citep{gbn} & CNN-FCF & \citep{cnn-fcf} \\
					SFP & \citep{sfp}& LEGR& \citep{legr} & FPGM & \citep{fpgm} \\
					TAS & \citep{tas}& GS& \citep{GS} & AutoPruner & \citep{autopruner} \\
					CGNet & \citep{cgnet}& MobileNetV2& \citep{mobilenetv2} & NA & \citep{na} \\
					MFP & \citep{mfp}& AMC& \citep{amc} & DMCP & \citep{dmcp} \\
					MuffNet & \citep{muffnet}& DCP& \citep{dcp} & FBS & \citep{fbs} \\
					VGGNet & \citep{vgg}& GAL& \citep{gal} & Sliming & \citep{slimming} \\
					PS & \citep{ps}& AOFP& \citep{aofp} &  &  \\ \hline
			\end{tabular}
			\vspace{-2mm}
		\end{table*}

	\subsection{The efficiency of CafeNet in training one-shot supernet $\N$}   \label{a5}
	To investigate the searching efficiency of CafeNet, we examine the time cost of training 1 epoch with different radius $r$. As shown in Table \ref{Efficiency},  the training time of CafeNet raised with an increase of $r$, which is because $r$ represents the preset allowed offset for free channels, thus increasing the times of forwarding for each batch of information. However, since CafeNet involves multi-times forward and once backward and the time cost of forwarding is much shorter than the backward, the overall training efficiency will not be greatly affected. To balance the performance and search efficiency, we set $r=1$ for all experiments with CafeNet.
	
		\begin{table*}[h]
		\centering
		\small
		\caption{The efficiency of CafeNet in training one-shot  supernet $\N$.}
		\label{Efficiency}
		\vspace{-2mm}
		{\begin{tabular}{c|c|c||c|c|c}
				\hline
				\multicolumn{3}{c||}{CIFAR-10} & \multicolumn{3}{c}{ImageNet} \\ \hline
				Local free&MobileNetV2&VGGNet&Local free&MobileNetV2& ResNet50  \\ \hline
				r=0 & 51.3s & 61.6s & r=0 & 428.4s & 748.1s \\
				r=1 & 61.1s & 72.4s & r=1 & 520.6s & 954.3s \\
				r=2 & 95.2s	& 109.7s& r=2 & 832.5s & 1569.6s \\
				r=3 & 203.6s& 258.4s& r=3 & 1872.3s & 3584.9s \\ \hline
				
		\end{tabular}}	
		\vspace{-2mm}
	\end{table*}

	\subsection{Performance gain of each part in CafeNet}   \label{a6}
	
	To investigate the performance gain of each part in CafeNet, we conduct experiments to search for 0.5$\times$ FLOPs MobileNetV2 and 0.5$\times$ FLOPs VGGNet on CIFAR-10 dataset and report their Top-1 accuracy. Since min-min optimization and max-max selection are always used simultaneously, for clarity, we abbreviate them as min-min optimization. As shown in Table \ref{CafeNet_analysis}, with evolutionary search, CafeNet can enjoy a performance gain of 0.91\% (0.74\%) accuracy on MobileNetV2 (VGGNet). Besides, similar results can also be obtained through the greedy search, which exactly shows that our CafeNet can boost to search for a decent network width.

	\begin{table}[h]
		\caption{Performance of searched 0.5$\times$ FLOPs MobileNetV2 and VGGNet on CIFAR-10 dataset with different supernets and searching methods.}
		\label{CafeNet_analysis}
		\centering
		\scriptsize
		\begin{tabular}{|c|c|c|c|c|c||c|c|} \hline
			\multicolumn{3}{|c|}{supernet} & \multicolumn{3}{c||}{searching}&  \multicolumn{2}{c|}{models} \\ \cline{1-8} 
			FLOPs-sensitive& Locally free & min-min& multi-stage & greedy & evolutionary& \multirow{2}*{MobileNetV2} & \multirow{2}*{VGGNet} \\ 
			bins &weight sharing pattern&optimization&search&search&search & &\\ \hline 
			& & & &\checkmark& & 94.27\% & 93.49\% \\
			\checkmark& & & &\checkmark& & 94.51\% & 93.65\% \\
			\checkmark&\checkmark&\checkmark& &\checkmark& & 94.76\% & 93.89\% \\  \hline
			& & & & &\checkmark& 94.53\% & 93.62\% \\
			\checkmark& & & & &\checkmark& 94.71\% & 93.74\% \\
			\checkmark&\checkmark&\checkmark& & &\checkmark& 95.03\% & 93.95\%\\
			\checkmark&\checkmark&\checkmark&\checkmark& &\checkmark& 95.44\% & 94.36\% \\ \hline
		\end{tabular} 
	\end{table}

	\subsection{Effect of FLOPs-sensitive bins}   \label{a7}
	Our proposed FLOPs-sensitive bins aim to reduce the search space by allowing the FLOPs to distribute more evenly over layers for the search unit. To explore the effect of FLOPs-sensitive bins as a search unit compared with the  uniform unit (uniform channel group), we search for network width under these two search units on CIFAR-10 dataset with MobileNetV2 and VGGNet, as shown in Table \ref{Experiments_grouping_supp}. In detail, compared with the uniform unit, higher accuracy under various FLOPs budgets can be achieved by leveraging FLOPs-sensitive bins.
	
	\begin{table*}[h]
		\centering
		\small
		\caption{Performance comparison of two forming search unit methods, \ie, baseline uniform groups, and  FLOPs-sensitive bins.}
		\label{Experiments_grouping_supp}
		\begin{tabular}{c|c|c||c|c|c}
			\hline
			\multicolumn{3}{c||}{VGGNet} & \multicolumn{3}{c}{MobileNetV2} \\ \hline
			FLOPs&Uniform groups&\textbf{FLOPs-sensitive bins}&FLOPs&Uniform groups&\textbf{FLOPs-sensitive bins}  \\ \hline
			189M& 94.12\% & \textbf{94.36\%} & 188M & 95.37\% & \textbf{95.56\%} \\ \hline
			154M& 93.87\% & \textbf{94.23\%} & 144M & 95.12\% & \textbf{95.44\%} \\ \hline
			115M& 93.64\% & \textbf{94.01\%} & 44M & 94.99\% & \textbf{95.31\%} \\ \hline
			76M& 93.25\% & \textbf{93.67\%} & 28M & 93.62\% & \textbf{94.11\%} \\ \hline
		\end{tabular}	
	\end{table*}

	\newpage
	
	\subsection{More detailed results of ImageNet dataset for Table \ref{Experiments_Imagenet}}   \label{a8}
	To explore the effectiveness of CafeNet, we also implement our algorithm on ResNet34 and ResNet18 with the same training strategy as ResNet50. The original ResNet34 and ResNet18 has 21.8M, 11.7M and 3.6G, 1.8G FLOPs with 74.9\%, 71.5\% Top-1 accuracy, respectively. The detailed results are reported in Table \ref{Experiments_Imagenet_exp}, our 0.75$\times$ FLOPs ResNet34 and ResNet18 can achieve close performance to the origin model. Besides, under the tiny FLOPs budget (\ie, 10\% FLOPs), CafeNet can outperform the unform baseline by a large margin.

	\begin{table*}[h]
		\centering
		\scriptsize
		\caption{Performance comparison of ResNet50, ResNet34, ResNet18 and MobileNetV2 on ImageNet. References of baseline methods are summarized in appendix \ref{a4}.}
		\label{Experiments_Imagenet_exp}
		\vspace{-2mm}
		\begin{tabular}{c|l|p{0.5cm}<{\centering}p{0.5cm}<{\centering}|p{0.6cm}>{\centering}p{0.6cm}<{\centering}||c|l|p{0.5cm}<{\centering}p{0.5cm}<{\centering}|p{0.6cm}<{\centering}p{0.6cm}<{\centering}}\hline
			\multicolumn{6}{c||}{ResNet50} & \multicolumn{6}{c}{ResNet34}\\ \hline
			&Methods&FLOPs&Param&Top-1&Top-5&&Methods&FLOPs&Param&Top-1&Top-5\\ \hline
			\multirow{8}{*}{3G} &AutoSlim & 3.0G & 23.1M & 76.0\% & -  & \multirow{10}*{2.7G} & Rethinking	 & 2.79G & - & 72.9\% & -  \\
			& MetaPruning & 3.0G & - & 76.2\% & - && MIL & 2.75G & - & 73.0\% & - \\
			& LEGR & 3.0G & -  & 76.2\% & - && PF & 2.79G & -  & 72.1\% & - \\
			& Uniform & 3.0G & 19.1M & 75.9\% & 93.0\% && Uniform & 2.7G & - & 72.2\% & 90.9\% \\
			& Random & 3.0G & - & 75.2\% & 92.5\% && Random & 2.7G & - & 71.6\% & 90.8\% \\
			& \textbf{CafeNet-R} & 3.0G & 22.6M & \textbf{77.1\%} & 94.3\% && \textbf{CafeNet-R} & 2.7G & 18.6M & \textbf{74.4\%} & 92.0\% \\ 
			& \textbf{CafeNet-E} & 3.0G & 23.8M & \textbf{77.4\%} & 94.5\% && \textbf{CafeNet-E} & 2.7G & 20.4M & \textbf{74.8\%} & 92.3\% \\ \cline{1-6}
			
			\multirow{14}*{2G} & GBN & 2.4G & 31.8M & 76.2\% & 92.8\% && CNN-FCF & 2.7G & 15.9M & 73.6\% & 91.5\% \\  
			& SFP & 2.4G & - & 74.6\% & 92.1\% && \textbf{CafeNet-R} & 2.5G & 19.3M & \textbf{74.0\%} & 91.9\% \\
			& LEGR & 2.4G & - & 75.7\% & 92.7\% && \textbf{CafeNet-E} & 2.5G & 20.2M & \textbf{74.5\%} & 92.1\% \\ \cline{7-12}
			& FPGM & 2.4G & - & 75.6\% & 92.6\% &\multirow{10}*{1.8G} & FPGM & 2.2G & - & 72.5\% & - \\
			& TAS & 2.3G & - & 76.2\% & 93.1\% && SFP & 2.2G & - & 71.8\% & 90.3\% \\
			& MetaPruning & 2.0G & -
			& 75.4\% & - && CNN-FCF& 2.2G & 12.6M & 72.8\% & 91.0\% \\
			& AutoSlim & 2.0G & 20.6M & 75.6\% & - && GS & 2.1G & - & 72.9\% & - \\
			& Uniform & 2.0G & 13.3M & 75.1\% & 92.7\% && Uniform & 1.8G & - & 71.6\% & 90.3\% \\
			& Random & 2.0G & - & 74.6\% & 92.2\% && Random & 1.8G & - & 71.1\% & 89.9\% \\
			& \textbf{CafeNet-R} & 2.0G & 19.1M & \textbf{76.5\%} & 93.1\% && \textbf{CafeNet-R} & 1.8G & 17.2M & \textbf{73.1\%} & 91.4\% \\
			& \textbf{CafeNet-E} & 2.0G & 18.4M & \textbf{76.9\%} & 93.3\% && \textbf{CafeNet-E} & 1.8G & 16.9M & \textbf{73.4\%} & 91.5\% \\ \cline{1-6}

			\multirow{8}*{1G} & AutoPruner & 1.4G & -  & 73.1\% & 91.3\% && CGNet & 1.8G & - & 71.3\% & - \\
			& MetaPruning & 1.0G & - & 73.4\% & - && CNN-FCF & 1.7G & 9.6M & 71.3\% & 90.2\% \\  \cline{7-12}
			& AutoSlim & 1.0G & - & 74.0\% & - &\multirow{10}*{1G-}& CNN-FCF & 1.2G & 7.1M & 69.7\% & 89.3\% \\
			& Uniform & 1.0G & 6.6M& 73.1\% & 91.8\% && CGNet & 1.2G & - & 70.2\% & -\\
			
			& Random & 1.0G & - & 72.2\% & 91.4\% && Uniform & 0.9G & - & 69.5\% & 89.4\% \\
			& \textbf{CafeNet-R} & 1.0G & 11.2M & \textbf{74.9\%} & 92.3\% && Random & 0.9G & - & 69.1\% & 88.9\% \\
			& \textbf{CafeNet-E} & 1.0G & 12M & \textbf{75.3\%} & 92.6\% && \textbf{CafeNet-R} & 0.9G & 10.1M & \textbf{71.8\%} & 89.5\% \\ \cline{1-6}
			\multirow{5}*{570M} & AutoSlim & 570M & - & 72.2\% & - && \textbf{CafeNet-E} & 0.9G & 9.8M & \textbf{72.1\%} & 89.8\% \\ 
			& Uniform & 570M & 4.0M & 71.6\% & 90.6\% && Uniform & 0.36G & - & 59.9\% & 82.3\% \\
			& Random & 570M & - & 69.4\% & 90.3\% && Random & 0.36G & - & 56.2\% & 80.6\% \\
			& \textbf{CafeNet-R} & 570M & 11.3M & \textbf{72.7\%} & 90.9\% && \textbf{CafeNet-R} & 0.36G & 3.5M & \textbf{63.3\%} & 85.2\% \\
			& \textbf{CafeNet-E} & 570M & 12.0M & \textbf{73.3\%} & 91.2\% && \textbf{CafeNet-E} & 0.36G & 3.6M & \textbf{64.0\%} & 85.4\% \\ \hline 
			\hline
			
			\multicolumn{6}{c||}{MobileNetV2} & \multicolumn{6}{c}{ResNet18}\\ \hline
			&Methods&FLOPs&Param&Top-1&Top-5&&Methods&FLOPs&Param&Top-1&Top-5\\ \hline
			\multirow{7}*{200M} & MetaPruning & 217M & - & 71.2\% & - & \multirow{6}*{1.2G} & TAS  & 1.2G & - & 69.2\% & 89.2\% \\
			& LEGR & 210M & - & 71.4\% & - && MIL & 1.2G & - & 66.3\% & 86.9\% \\
			& AutoSlim & 207M & 4.1M & 73.0\% & - && Uniform  & 1.2G & 8.5M & 68.8\% & 88.5\%  \\ 
			
			& Uniform & 217M & 2.7M & 70.9\% & 89.4\% && Random & 1.2G & -& 68.4\% & 88.1\% \\
			& Random & 217M & - & 70.3\% & 89.1\% && \textbf{CafeNet-R} & 1.2G & 11.5M & \textbf{70.8\%} & 89.8\% \\
			& \textbf{CafeNet-R} & 217M & 3.0M & \textbf{73.3\%} & 91.1\% && \textbf{CafeNet-E} & 1.2G & 11.3M & \textbf{71.2\%}& 89.9\% \\ \cline{7-12}
			& \textbf{CafeNet-E} & 217M & 3.3M & \textbf{73.4\%} & 91.2\% &\multirow{13}*{1G}& NA & 1.17G & - & 69.4\% & 88.71\% \\ \cline{1-6}

			\multirow{9}*{150M} & LEGR & 150M & - & 70.8\% & - && SFP & 1.05G & - & 67.1\% & 87.8\% \\
			& TAS & 150M & - & 70.9\% & - && MFP & 1.05G & - & 68.3\% & 88.3\% \\
			& AMC & 150M & - & 70.8\% & - && DMCP & 1.04G & - & 69.2\% & - \\
			& LEGR & 150M & - & 69.4\% & - && FPGM & 1.04G & - & 68.4\% & 88.5\% \\
			& MuffNet & 149M & - & 63.7\% & - && DCP & 0.96G & - & 67.4\% & 87.6\% \\ 
			& Uniform & 150M & 2.0M & 69.3\% & 88.9\% && CGNet & 0.94G & - & 68.8\% & - \\
			& Random & 150M & - & 68.8\% & 88.7\% && MFP& 0.9G & - & 67.1\% & 87.5\% \\ 
			
			& \textbf{CafeNet-R} & 150M & 2.7M & \textbf{71.9\%} & 90.0\% && FBS & 0.9G & - & 68.2\% & 88.2\% \\
			& \textbf{CafeNet-E} & 150M & 3.0M & \textbf{72.4\%} & 90.4\% && Uniform & 0.9G & 6.0M & 67.1\% & 87.5\% \\ \cline{1-6}
			
			\multirow{11}*{100M} & MetaPruning & 105M & - & 65.0\% & - && Random & 0.9G & - & 66.7\% & 87.1\% \\
			& Uniform & 105M & 1.5M & 65.1\% & 89.6\% && \textbf{CafeNet-R} & 0.9G & 9.7M & \textbf{69.6\%} & 88.8\% \\
			& Random & 105M & - & 63.9\% & 89.2\% && \textbf{CafeNet-E} & 0.9G & 10.2M & \textbf{69.8\%} & 89.0\% \\ \cline{7-12}
			& \textbf{CafeNet-R} & 106M & 2.2M & \textbf{68.2\%} & 88.2\% & \multirow{8}*{0.45G-} & Uniform & 450M & 2.9M & 61.6\% & 83.6\% \\
			& \textbf{CafeNet-E} & 106M & 2.1M & \textbf{68.7\%} & 88.5\% && Random & 450M & - & 59.8\% & 82.3\% \\ 
			& MuffNet & 50M & - & 50.3\% & - && \textbf{CafeNet-R} & 450M & 5.0M & \textbf{65.2\%} & 86.0\% \\
			& MetaPruning & 43M & -& 58.3\% & - && \textbf{CafeNet-E} & 450M & 5.8M & \textbf{65.6\%} & 86.2\% \\
			
			& Uniform & 50M & 0.9M & 59.7\% & 82.0\% && Uniform & 180M & 1.1M & 53.7\% & 77.5\% \\
			& Random & 50M & - & 57.4\% & 81.2\% && Random & 180M & - & 51.6\% & 76.9\% \\
			& \textbf{CafeNet-R} & 50M & 1.7M & \textbf{64.3\%} & 85.2\% && \textbf{CafeNet-R} & 180M & 1.9M & \textbf{57.8\%} & 81.7\% \\
			& \textbf{CafeNet-E} & 50M & 1.6M & \textbf{64.9\%} & 85.4\% && \textbf{CafeNet-E} & 180M & 2.0M & \textbf{58.4\%} & 81.9\% \\     \cline{1-6} \cline{7-12}
		\end{tabular}	
		\vspace{-0.6cm}
	\end{table*}

	\subsection{More detailed results of CIFAR-10 and CelebA dataset for Table \ref{Experiments_CelebA} and \ref{Experiments_COCO}}   \label{a9}
	
	To examine the hyperparameters setting with CIFAR-10 and CelebA dataset, we also explore the searching results from two vanilla baselines named Random and Uniform, as described in Section 4. The detailed results on CIFAR-10 dataset and CelebA dataset are reported in Table \ref{Experiments_CIFAR_supp} and Table \ref{Experiments_CelebA_supp}, respectively. Besides, for CelebA dataset, the layer widths of fully-connection(FC) layers are not implemented to be searched and keep the same as \citep{mt}, since they only takes a tiny amount of FLOPs to the network.
	\begin{table*}[h]
		\centering
		\scriptsize
		\caption{Performance comparison of MobileNetV2 and VGGNet on CIFAR-10 dataset.}
		\label{Experiments_CIFAR_supp}
		\vspace{-3mm}
		{\begin{tabular}{c|l|cc|c||c|l|cc|c}
				\hline
				\multicolumn{5}{c||}{MobileNetV2} & \multicolumn{5}{c}{VGGNet} \\ \hline
				Groups&Methods&FLOPs&Parameters&Accuracy&Groups&Methods&FLOPs&Parameters&Accuracy  \\ \hline
				\multirow{5}*{200M} & DCP & 218M & 1.7M & 94.75\% &\multirow{10}*{200M} & GAL & 190M & - & 93.80\% \\
				& Uniform & 188M & 1.4M & 94.57\%&& DCP & 199M & 10.4M & 94.16\% \\
				& Random & 188M & - & 94.20\% && Sliming & 199M & 10.4M & 93.80\% \\
				& \textbf{CafeNet-R} & 188M & 1.4M & \textbf{95.44\%} && Uniform & 189M & 9.5M & 93.37\% \\ 
				& \textbf{CafeNet-E} & 188M & 1.5M & \textbf{95.56\%} && Random & 189M & - & 93.06\% \\ \cline{1-5}
				
				\multirow{5}*{144M} & MuffNet & 175M & - & 94.71\% && \textbf{CafeNet-R} & 189M & 8.3M & \textbf{94.27\%} \\ 
				& Uniform & 144M & 1.1M & 94.28\% && \textbf{CafeNet-E} & 189M & 8.0M & \textbf{94.36\%} \\ \cline{6-10}
				& Random & 144M & - & 93.76\% &\multirow{13}*{100M+} & PS & 156M & - & 93.63\% \\
				& \textbf{CafeNet-R} & 144M & 1.2M & \textbf{95.28\%} && Uniform & 154M & 7.7M & 93.11\% \\
				& \textbf{CafeNet-E} & 144M & 1.1M & \textbf{95.44\%} && Random & 154M & - & 92.86\% \\ \cline{1-5}
				
				\multirow{5}*{44M} & AutoSlim & 88M & 1.5M & 93.20\% && \textbf{CafeNet-R} & 154M & 3.4M & \textbf{94.09\%} \\
				& AutoSlim & 59M & 0.7M & 93.00\% && \textbf{CafeNet-E} & 154M & 3.1M & \textbf{94.23\%} \\
				& MuffNet & 45M & - & 93.12\% && AOFP & 124M & - & 93.84\% \\	
				& Uniform & 44M & 0.3M & 92.72\% && Uniform & 115M & 5.9M & 92.95\% \\
				& Random & 44M & - & 92.24\% && Random & 115M & - & 92.77\% \\
				& \textbf{CafeNet-R} & 44M & 0.4M & \textbf{95.16\%} && \textbf{CafeNet-R} & 115M & 2.4M & \textbf{93.87\%} \\  
				& \textbf{CafeNet-E} & 44M & 0.4M & \textbf{95.31\%} && \textbf{CafeNet-E} & 115M & 2.1M & \textbf{94.01\%} \\ \cline{1-5}	 \cline{6-10}
				
				\multirow{5}*{28M} & AutoSlim & 28M & 0.3M & 92.00\% & \multirow{5}*{76M} & CGNets & 92M & - & 92.88\% \\
				& Uniform & 28M & 0.3M & 91.87\% && Uniform & 76M & 3.9M & 92.32\% \\
				& Random & 28M & - & 91.36\% && Random & 76M & - & 91.67\% \\
				& \textbf{CafeNet-R} & 28M & 0.2M & \textbf{93.87\%} && \textbf{CafeNet-R} & 76M & 2.2M & \textbf{93.36\%} \\ 
				& \textbf{CafeNet-E} & 28M & 0.2M & \textbf{94.11\%} && \textbf{CafeNet-E} & 76M & 1.4M & \textbf{93.67\%} \\ \hline
				
		\end{tabular}}	
		\vspace{-4mm}
	\end{table*}

	\begin{table*}[h]
		\centering
			\scriptsize
		\caption{Performance comparison of MobileNetV2 and ResNet18 on CelebA dataset.}
		\label{Experiments_CelebA_supp}
		{\begin{tabular}{l|lc|c||l|lc|c}
				\hline
				\multicolumn{4}{c||}{MobileNetV2} & \multicolumn{4}{c}{ResNet18} \\ \hline
				Methods&FLOPs&Parameters&Accuracy&Methods&FLOPs&Parameters&Accuracy  \\ \hline
				Uniform & 162M & - & 91.97\% & Uniform & 1G & - & 92.03\% \\
				Random & 162M & - & 91.76\% & Random & 1G & - & 91.84\% \\
				\textbf{CafeNet-R} & 162M & 1.7M & \textbf{92.12\%}& \textbf{CafeNet-R} & 1G & 1.7M & \textbf{92.17\%} \\ 
				\textbf{CafeNet-E} & 162M & 1.8M & \textbf{92.19\%} & \textbf{CafeNet-E} & 1G & 1.8M & \textbf{92.25\%} \\ \hline

				Uniform & 106M & - & 91.92\% & Uniform & 619M & - & 91.93\% \\
				Random & 106M & - & 91.63\% & Random & 619M & - & 91.67\% \\
				\textbf{CafeNet-R} & 106M & 1.2M & \textbf{92.09\%} & \textbf{CafeNet-R} & 619M & 0.4M & \textbf{92.13\%} \\ 
				\textbf{CafeNet-E} & 106M & 1.2M & \textbf{92.16\%} & \textbf{CafeNet-E} & 619M & 0.5M & \textbf{92.18\%} \\ \hline
				
				Uniform & 51M & - & 91.73\% & Uniform & 316M & - & 91.79\% \\
				Random & 51M & - & 91.52\% & Random & 316M & - & 91.62\% \\
				\textbf{CafeNet-R} & 51M & 0.6M & \textbf{92.03\%} & \textbf{CafeNet-R} & 316M & 3.1M & \textbf{92.07\%} \\ 
				\textbf{CafeNet-E} & 51M & 0.5M & \textbf{92.13\%} & \textbf{CafeNet-E} & 316M & 3.4M & \textbf{92.16\%} \\ \hline
				
				Uniform & 21M & - & 91.63\% & Uniform & 130M & - & 91.66\%  \\
				Random & 21M & - & 91.42\% & Random & 130M & - & 91.51\% \\
				\textbf{CafeNet-R} & 21M & 0.2M & \textbf{91.71\%} & \textbf{CafeNet-R} & 130M & 1.2M & \textbf{91.83\%} \\ 
				\textbf{CafeNet-E} & 21M & 0.2M & \textbf{91.85\%} & \textbf{CafeNet-E} & 130M & 1.0M & \textbf{91.92\%} \\ \hline
				
		\end{tabular}}	
		\vspace{-2mm}
	\end{table*}
	
	\newpage
	
\begin{figure}[h]
		\vspace{-2mm}
		\centering
		\subfigure[Performance of 50\% FLOPs MobileNetV2 \wrt 40 labels on CelebA dataset.]{
			\label{fig:celeba:a} 
			\includegraphics[width=0.95\linewidth]{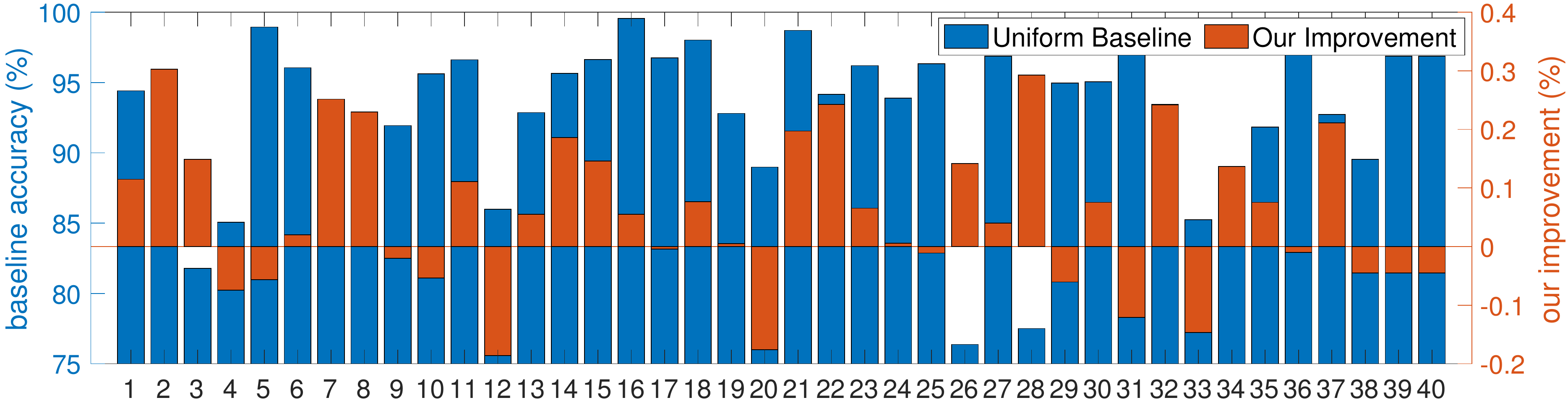}}
		\subfigure[Performance of 10\% FLOPs MobileNetV2 \wrt 40 labels on CelebA dataset.]{
			\label{fig:celeba:b}
			\includegraphics[width=0.95\linewidth]{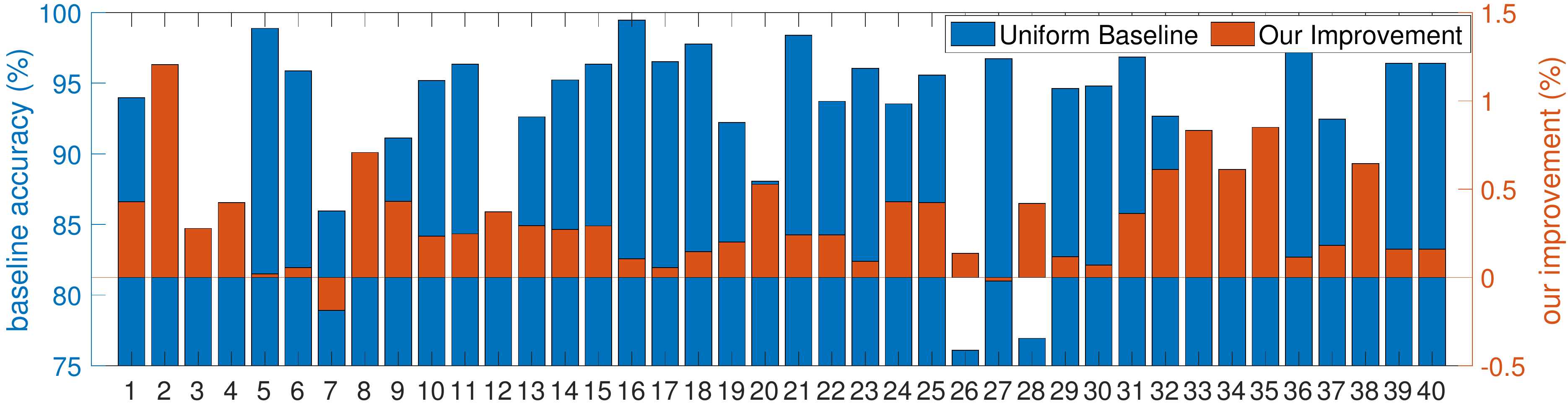}}
		\vspace{-2mm}
		\caption{Performance of 50\% MobileNetV2 \wrt 40 labels on CelebA dataset. The blue bar refers to the performance of network width searched by CafeNet, while the red bar indicates the performance gap in comparison to the uniform baseline.}
		\label{CelebA_40}
		\vspace{-5mm}
	\end{figure}

\textbf{Classification results for all 40 labels on CelebA dataset}. As shown in Fig.\ref{CelebA_40}, our searched network width has better performance in comparing to the uniform baseline, especially for small FLOPs budget and labels that are difficult to be classified.

	\newpage
	\subsection{More experimental results with aligned training recipes of AutoSlim \cite{autoslim}}   \label{a_supp}
	To examine the performance of our searched network width, we retrain the searched width of MobileNetV2 and ResNet50 with aligned training recipes of AutoSlim \cite{autoslim}. In detail, we report the retraining results of width with similar FLOPs to AutoSlim in Table \ref{Experiments_aligned}. 
	
	\begin{table*}[h]
	\centering
	\caption{Performance of MobileNetV2 and ResNet50 with aligned training recipes of AutoSlim \cite{autoslim}.}
	\label{Experiments_aligned}
		{\begin{tabular}{l|lc|c|c}
				\hline
				\multicolumn{5}{c}{ResNet50} \\ \hline
				Methods&FLOPs&Parameters&Top-1&Top-5 \\ \hline
				AutoSlim & 3.0G & 23.1M & 76.0\% & - \\ 
				CafeNet-R & 3.0G & 22.6M & 76.2\% &93.1\% \\ 
				CafeNet-E & 3.0G & 23.8M & 76.4\% &93.3\% \\ \hline
				AutoSlim & 1G & - & 74.0\% & - \\ 
				CafeNet-R & 1G & 11.2M & 74.2\% &91.9\% \\ 
				CafeNet-E & 1G & 12M & 74.5\% &92.1\% \\  \hline
				AutoSlim & 570M & - & 72.2\% & - \\
				CafeNet-R & 570M & 11.3M & 72.1\% &90.7\% \\ 
				CafeNet-E & 570M & 12M & 72.6\% &91.0\% \\ \hline
				\multicolumn{5}{c}{MobileNetV2} \\  \hline
				AutoSlim & 207M & 4.1M & 73.0\%& - \\ 
				CafeNet-R & 217M & 3.0M & 73.1\%& 90.9\% \\ 
				CafeNet-E & 217M & 3.3M & 73.2\%&91.1\% \\  \hline
		\end{tabular}}	
	\end{table*}

As Table \ref{Experiments_aligned} shows, with similar FLOPs and the same training recipes, CafeNet can search the width with higher accuracy, which proves the effectiveness of our method.

	\newpage
	\subsection{Visualization of searched network width}   \label{a10}
	
\begin{figure}[h]
		\centering
		\subfigure[Network width of searched EfficientNet-B0 on ImageNet dataset.]{
			\label{fig:vis:c}
			\includegraphics[width=0.94\linewidth]{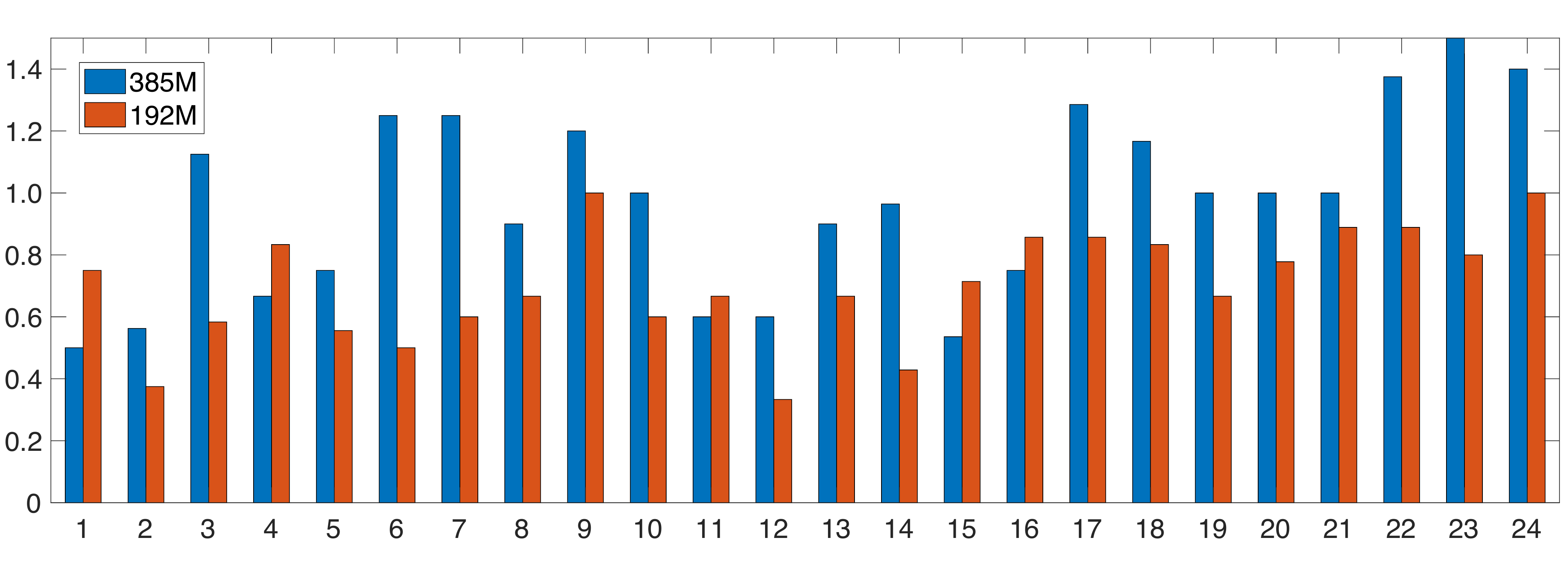}}
		\subfigure[Network width of searched MobileNetV2 on ImageNet dataset.]{			
			\label{fig:vis:a} 
			\centering
			\includegraphics[width=0.96\linewidth]{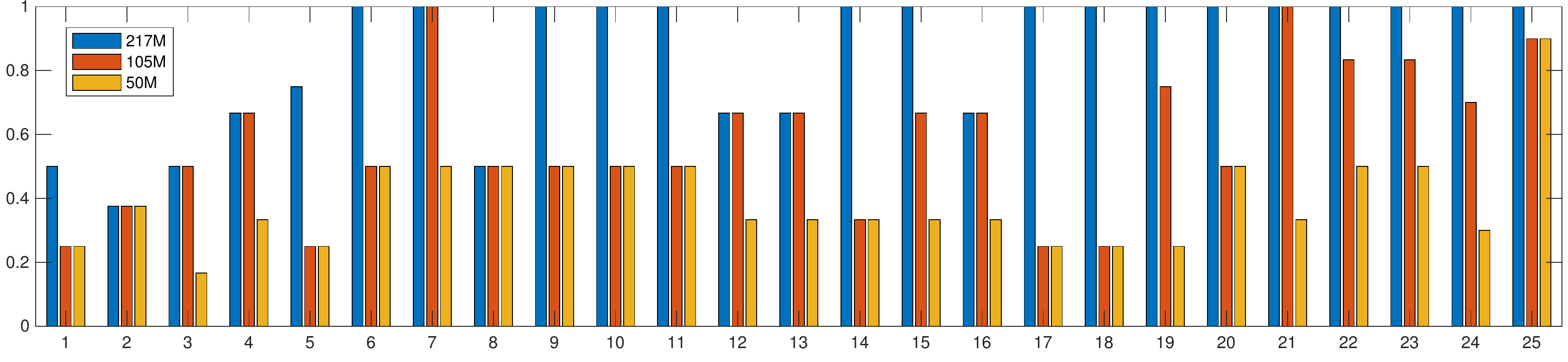}}
		\subfigure[Network width of searched ResNet50 on ImageNet dataset.]{
			\label{fig:vis:b}
			\includegraphics[width=0.96\linewidth]{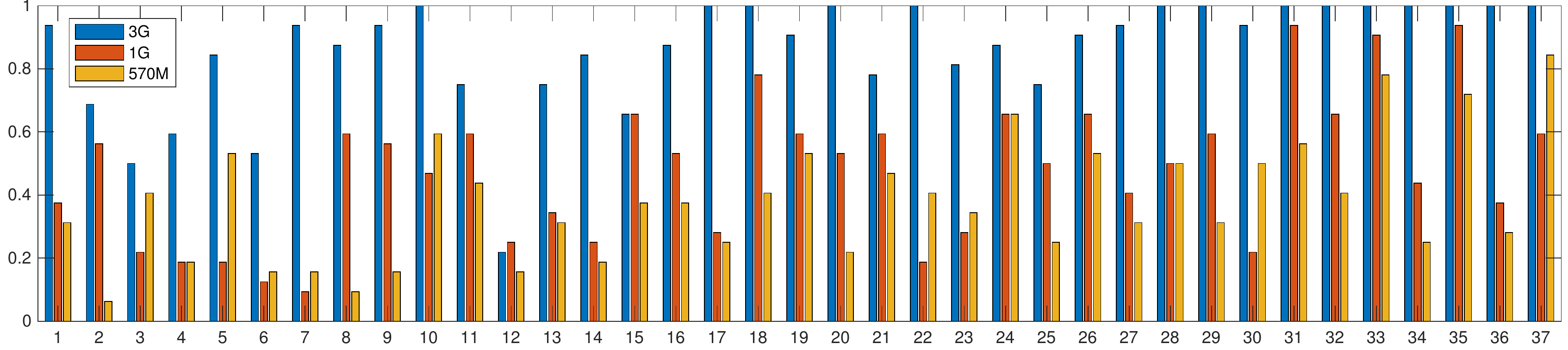}}
		\vspace{-3mm}
		\caption{Visualization of Pruned models.}
		\label{vis}
	\end{figure}

	For intuitively understanding, we visualize our searched networks with ImageNet dataset, as shown in Fig.\ref{vis}. For clarity, we show the retained ratio of network width from the original model. Note that, EfficientNet-B0, ResNet50, and MobileNetV2 have SE block, skipping, or depthwise layers; we merged these layers, which are required to have the same network width for visual clarity. For EfficientNet-B0, we also did not show the channel of the SE block because these layers are fixed during the network width search.
	
	For ImageNet dataset, from Fig.\ref{vis}, we can see that layer width in skipping layers are more inclined to be preserved, which means skipping layers are very useful for classification. Besides, with a large FLOPs budget, width in the 1$\times$ 1 convolution layers shrink more than in 3$\times$ 3 convolution layers, which means 1$\times$ 1 convolution may contribute less to classification performance. However, we observe an opposite phenomenon for width pruned from small FLOPs budgets, which implies the network is forced to use more 1$\times$ 1 convolution layers to extract information from feature maps instead of 3$\times$ 3 convolution layers.
	
	\newpage

\subsection{Visualization of bins}   \label{a11}
\begin{figure}[h]
		\centering
		\subfigure[EfficientNet-B0]{
			\label{fig:EfficientNet-b0}
			\includegraphics[width=1.0\linewidth]{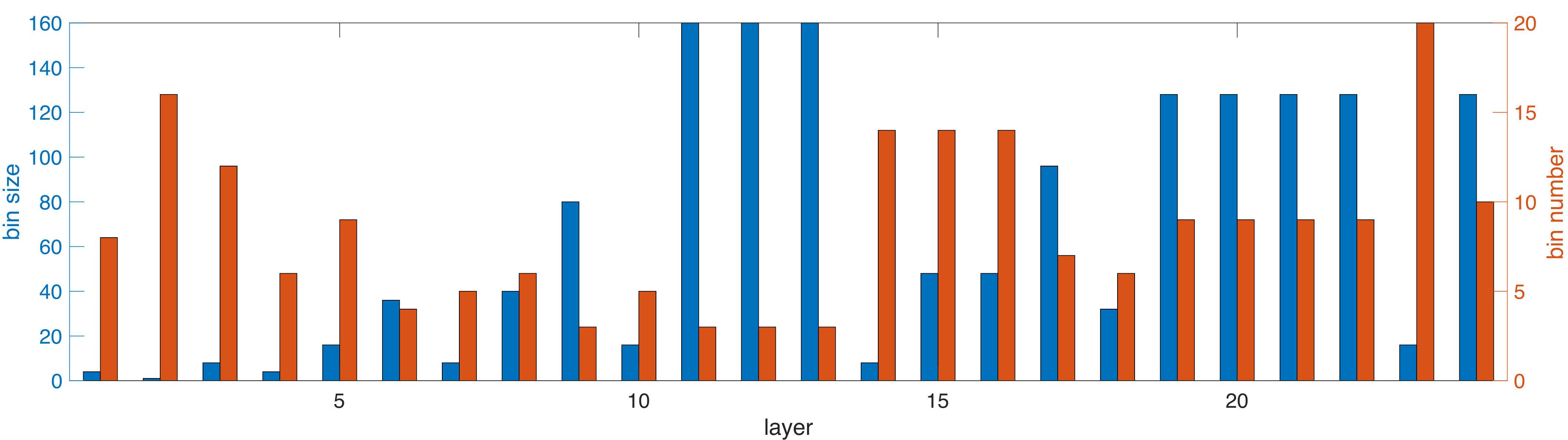}
		}
		\subfigure[MobileNetV2]{
			\label{fig:MobileNetV2} 
			\centering
			\includegraphics[width=1.0\linewidth]{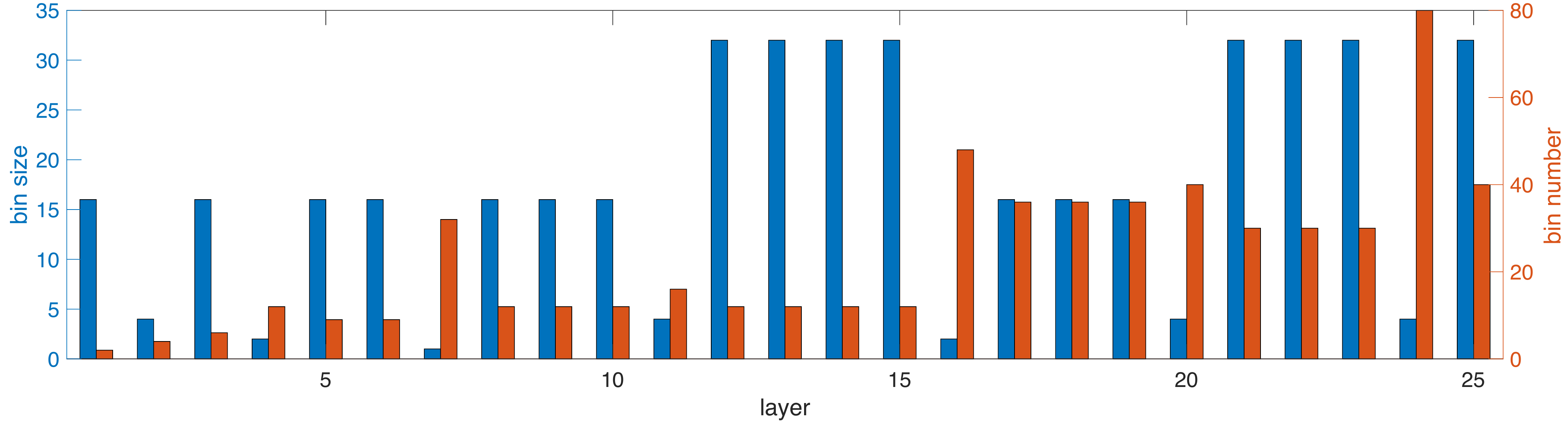}}
		\vspace{-3mm}
		\caption{Visualization of bin size and bin number of CafeNet. } \label{bins}
		\vspace{-5mm}
	\end{figure}

For intuitively understanding of FLOPs-sensitive bins, as shown in Fig.\ref{bins}, we visualize the number of channels contained in each bin (\blue{bin size}) and the number of bins in each layer (\red{bin number}) with MobileNetV2 and VGGNet. Similar to the visualization of the searched network width, we merged the SE blocks, skipping, and depthwise layers for clarity. To make the bins from all layers have a similar contribution to FLOPs, the number of bins varies significantly between layers. For EfficientNet-B0, the size of bin changes according to the different selected blocks (\ie, blocks with 3$\times$3 or 5$\times$5 convolution kernel size), \ie, the convolution of 3$\times$3 or 5$\times$5, and the size of bin becomes larger at the middle and the end of the network. For MobileNetV2, since only 3$\times$3 convolution is involved, the bin size changes more regularly, and the number of bins gradually increases from the front to the end of the network.

	\subsection{Effect of min-min optimization in Eq.\eqref{minmin0}\textasciitilde\eqref{maxmax}}   \label{a12}
	
\begin{figure}[h]
		\centering
		\subfigure[Histogram of Top-1 accuracy by evolutionary with MobileNetV2.]{
			\label{fig:samp:a} 
			\centering
			\includegraphics[width=0.48\linewidth]{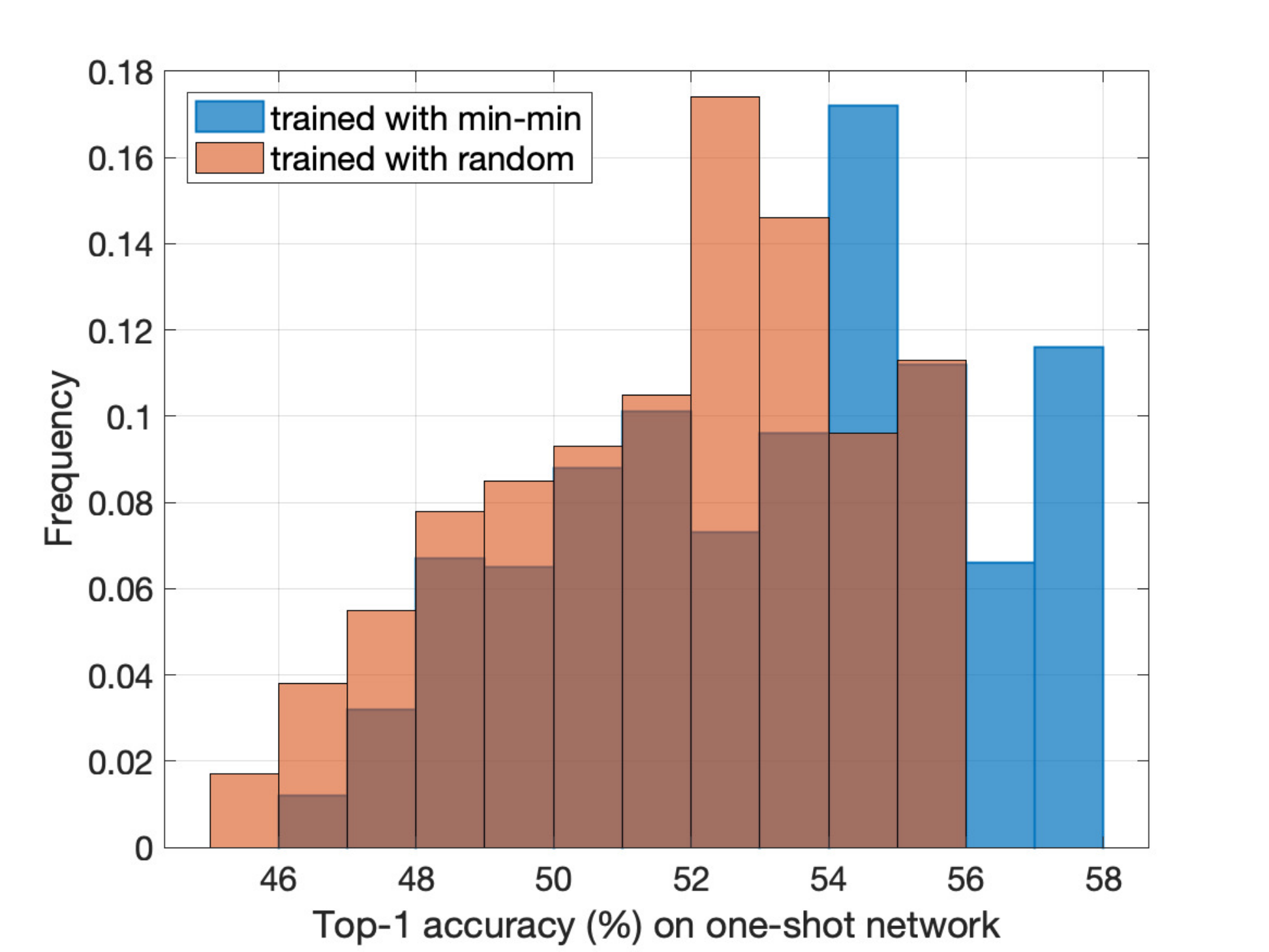}}
		\subfigure[Histogram of Top-1 accuracy by evolutionary with ResNet50.]{
			\label{fig:samp:b}
			\includegraphics[width=0.48\linewidth]{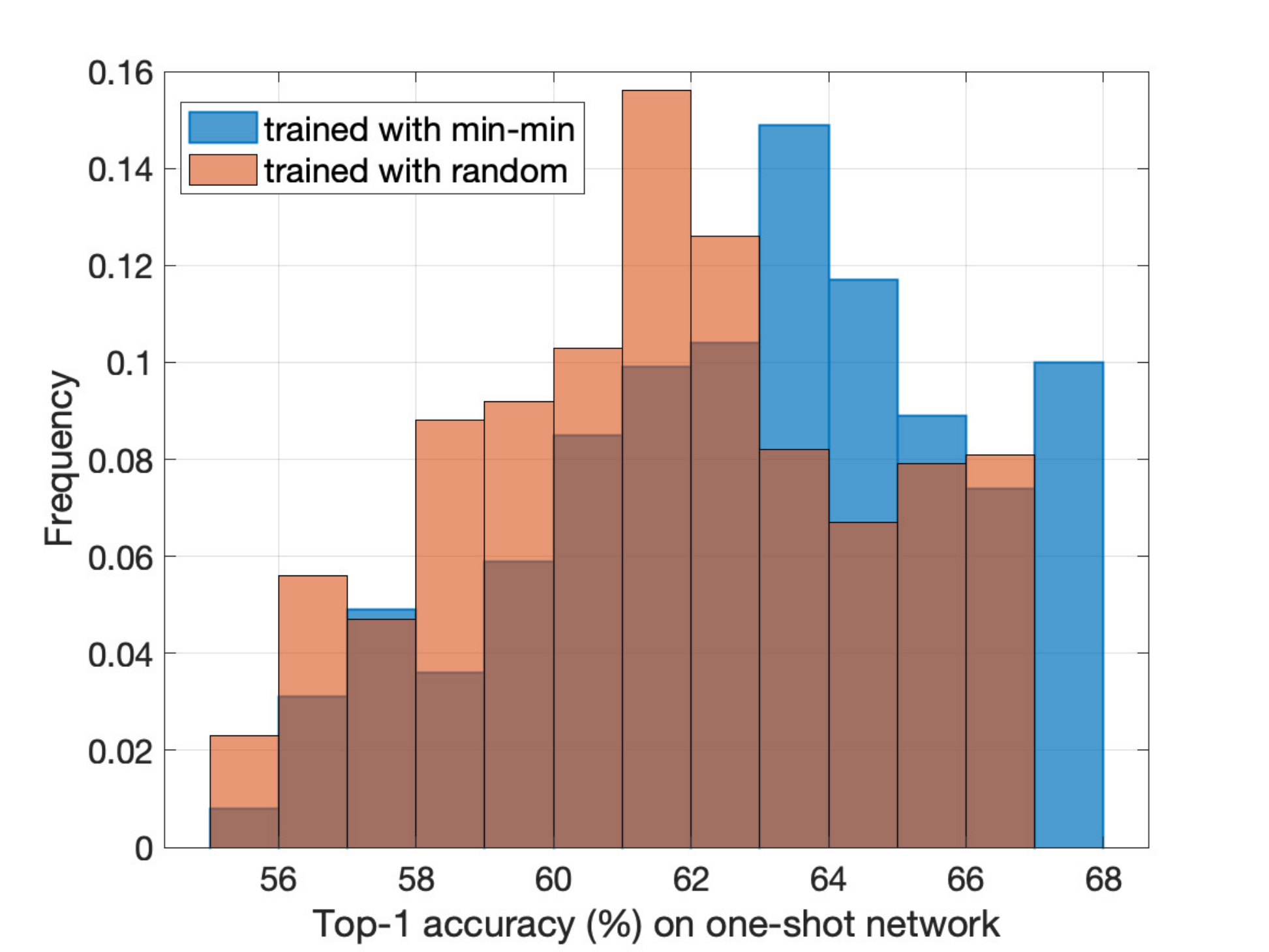}}
		\vspace{-3mm}
		\caption{Histogram of Top-1 accuracy of network width trained from min-min optimization and random optimization by evolutionary \wrt~MobileNetV2 and ResNet50. }
		\label{samp}
		\vspace{-5mm}
	\end{figure}

	To train the supernet, we investigate random optimization and min-min optimization with 50\% FLOPs MobileNetV2 and ResNet50 on ImageNet dataset, for instance. In detail, based on the trained supernet, we use the evolution search to obtain the optimized network width and record 1000 network width for both random optimization and min-min optimization, and report the results by a statistical distribution histogram, as shown in Fig.\ref{samp}. During sampling, the average performance of network width with our min-min optimization achieved 62.4\% (52.7\%) Top-1 accuracy with ResNet50 (MobileNetV2), which is 1.4\% (1.3\%) better than results from random optimization. In addition, width with the best performance is trained from scratch for evolution, as shown in Table \ref{Experiments_sampling}.
	
	\begin{table*}[h]
		\centering
		\caption{Performance comparison of weights updating methods.}
		\label{Experiments_sampling}
		\begin{tabular}{c|c|c||c|c|c}
			\hline
			\multicolumn{3}{c||}{0.5$\times$ResNet50} & \multicolumn{3}{c}{0.5$\times$MobileNetV2} \\ \hline
			Method&Sampling&CafeNet&Method&Sampling&CafeNet  \\ \hline
			Random & 67.14\% & 76.34\% & Random & 56.22\%  & 71.97\% \\ \hline
			\textbf{min-min} & \textbf{68.56\%} & \textbf{76.92\%} & \textbf{min-min} & \textbf{58.37\%} & \textbf{72.43\%} \\ \hline
			
		\end{tabular}	
	\end{table*}

	In Table \ref{Experiments_sampling}, we record the accuracy of Top-1 performance network width in supernet as Sampling, and than we training from scratch the width to report its Top-1 accuracy as CafeNet.

	\subsection{Effect of multi-stage search with evolving bins}   \label{a13}
	To examine the effect of multi-stage search with evolving bins, we report the Top-1 accuracy of MobileNetV2 on CIFAR-10 dataset with different training stages and the bin evolving speed $\alpha$ in Section 3.4, the results are summarized in Table  \ref{Experiments_evlutionbins_supp}. In detail, for CIFAR-10 dataset, our algorithm achieves superior performance with $\alpha$ set to 2 and searching in 3 stages, which indicates bin evolving strategy can lead to better results comparing with searching in one stage. However, when $\alpha$ is increased to 4, the performance of the searching results decreases a bit, which may be due to the larger search space that limits the ability of evolutionary search.

	\begin{table*}[h]
		\centering
		\caption{Performance of 50\% FLOPs MobileNetV2 and VGGNet on CIFAR-10 dataset with different training stages and bin evolving speed $\alpha$.}
		\label{Experiments_evlutionbins_supp}
		\begin{tabular}{c|c|c|c}
			\hline
			Number of training stages&Bin evolving speed $\alpha$&VGGNet&MobileNetV2  \\ \hline
			1& - & 93.95\% & 95.03\% \\ \hline
			2& 1 & 94.22\% & 95.31\% \\ \hline
			2& 2 & 94.29\% & 95.41\% \\ \hline
			2& 4 & 94.18\% & 95.29\% \\ \hline
			3& 1 & 94.29\% & 95.36\% \\ \hline
			3& 2 & 94.36\% & 95.44\% \\ \hline
			3& 4 & 94.21\% & 95.31\% \\ \hline
			
		\end{tabular}	
	\end{table*}

	\subsection{Estimation of the size of search space in multi-stage search}  \label{a14}
	
	Note that by leveraging the multi-stage searching strategy, we can further reduce the search space by following the searching "from coarse-grained to fine-grained" with evolving bins "from large to small". In this way, we directly investigate how the search space is reduced by the multi-stage searching strategy. We compare the search space of supernet under different training stages, bin size $\beta$, and bin evolving speed $\alpha$ in Table \ref{Experiments_searchspace_supp}. To fairly compare the results with different training stages, we set the size of the bin unit to be the same in Table \ref{Experiments_searchspace_supp}. And the search space is defined as the product of the number of bins in all layers.
	
	\begin{equation}
	\B^s_{space} =\sum_{j=1}^{s}\prod_{i=1}^{L}\frac{Channel_i^j}{b_{i}^j}
	\label{eq10}
	\end{equation}
	
	Where $Channel_i^j$ represents the total channels at layer $i$ in stage $j$, and $b_{i}^j$ indicates the bin size of layer $i$ in stage $j$. We examine the search space of 0.5$\times$ FLOPs MobileNetV2 on ImageNet dataset, as shown in Table \ref{Experiments_searchspace_supp}.

	\begin{table*}[h]
		\centering
		\caption{Search space of 0.5$\times$ FLOPs MobileNetV2 on ImageNet dataset.}
		\label{Experiments_searchspace_supp}
		\begin{tabular}{c|c|c|l|l}
			\hline
			\multicolumn{5}{c}{0.5$\times$ MobileNetV2} \\ \hline
			Stages&Bin size $\beta$&Evolve speed $\alpha$&Search space&Percent  \\ \hline
			1& - & - & 1.5$\times$ $10^{27}$ & 1 \\ 
			2& 1 & 2 & 2.0$\times$ $10^{19}$ & 1.3$\times$ $10^{-8}$ \\ 
			2& 2 & 4 & 4.8$\times$ $10^{10}$ & 3.0$\times$ $10^{-17}$\\ 
			3& 2 & 2 & 6.2$\times$ $10^{10}$ & 3.9$\times$ $10^{-17}$ \\ 
			4& 4 & 2 & 7.3$\times$ $10^{3}$  & 4.6$\times 10^{-24}$  \\ \hline

		\end{tabular}	
	\end{table*}

	In detail, our search space reduced by more than $1/(1.3\times 10^{-8})$ times from the original size (with one stage) by searching more than or equal to 2 stages. As a result, our proposed multi-stages search method can effectively reduce the search space.

	\subsection{Effect of smallest bin size $\beta$ in FLOPs-sensitive bins} \label{a15}
	As formulated in Eq.\eqref{eq7}, the total number of bins within each layer relies on the smallest bin size $\beta$. To examine the effect of $\beta$ \wrt~bin numbers and its corresponding performance, we perform experiments with MobileNetV2 on ImageNet dataset under 50\% FLOPs budget, as shown in Table \ref{Experiments_minbin_supp}. In detail, the performance of searched network width first increases when $\beta$ goes large but a little decreased after $\beta$ surpasses 0.5, which may result from that a very small $\beta$ induces a too huge search space thus is harmful to searching algorithms to find the optimized network width.
	
	\begin{table*}[h]
		\centering
		\caption{Performance of ResNet50 and MobileNetV2 on ImageNet dataset \wrt~different smallest bin size $\beta$ for FLOPs-sensitive bins.}
		\label{Experiments_minbin_supp}
		\begin{tabular}{c|c|c||c|c|c}
			\hline
			\multicolumn{3}{c||}{MobileNetV2} & \multicolumn{3}{c}{ResNet50} \\ \hline
			Smallest bin size $\beta$& Top-1 & Top-5 &Smallest bin size $\beta$& Top-1 & Top-5  \\ \hline
			4& 72.22\% & 90.22\% & 4 & 76.43\%& 93.01\% \\ \hline
			2& 72.32\% & 90.25\% & 2 & 76.73\%& 93.17\% \\ \hline
			1& 72.40\% & 90.38\% & 1 & 76.91\%& 93.33\% \\ \hline
			0.5& 72.34\% & 90.29\% & 0.5 & 76.82\%& 93.24\% \\ \hline
			0.25& 72.28\% & 90.31\% & 0.25 & 76.67\%& 93.14\%\\ \hline
			
		\end{tabular}	
	\end{table*}

\end{document}